\pdfoutput=1

\documentclass[11pt]{article}

\usepackage[]{emnlp2021}

\usepackage{times}
\usepackage{latexsym}

\usepackage[T1]{fontenc}

\usepackage[utf8]{inputenc}

\usepackage{microtype}

\usepackage{verbatim}
\usepackage{amssymb}
\usepackage{dsfont}
\usepackage{enumitem}
\usepackage{graphicx}
\usepackage{textcomp}
\usepackage{booktabs}
\usepackage{dirtytalk}
\usepackage{tikz}
\usepackage{url}
\usepackage{subcaption}
\usepackage{amsmath}
\usepackage{cleveref}
\newcommand\sect[1]{\S\ref{#1}}
\usepackage{multirow}
\usepackage{makecell}
\usepackage{colortbl}
\usepackage{xspace}
\usepackage{paralist}
\usepackage{stmaryrd}
\usepackage{footmisc}
\usepackage{amsmath}
\usepackage{microtype}

\usepackage{pgfplots}
\DeclareUnicodeCharacter{2212}{−}
\usepgfplotslibrary{groupplots,dateplot}
\usetikzlibrary{patterns,shapes.arrows}
\pgfplotsset{compat=newest}

\newcommand\cqa{CommonsenseQA\xspace}
\newcommand\itor{I$\rightarrow$R\xspace}
\newcommand\rtoo{R$\rightarrow$O\xspace}
\newcommand\itoor{I$\rightarrow$OR\xspace}
\newcommand\irtoo{IR$\rightarrow$O\xspace}

\usepackage{amsmath}
\newcommand\numberthis{\addtocounter{equation}{1}\tag{\theequation}}

\title{Measuring Association Between Labels and Free-Text Rationales}

\author{Sarah Wiegreffe\textsuperscript{$\clubsuit$} \hspace*{10mm} Ana Marasovi\'{c}\textsuperscript{$\dagger\Diamond$} \hspace*{10mm} Noah A. Smith\textsuperscript{$\dagger\Diamond$} \\ \\
  \textsuperscript{$\clubsuit$}School of Interactive Computing, Georgia Institute of Technology \\
    \textsuperscript{$\dagger$}Allen Institute for Artificial Intelligence\\
  \textsuperscript{$\Diamond$}Paul G. Allen School of Computer Science and Engineering, University of Washington\\
  \texttt{saw@gatech.edu, \{anam,noah\}@allenai.org} \\}

\begin{document}
\maketitle
\begin{abstract}
In interpretable NLP, we require faithful rationales that reflect the model's decision-making process for an explained instance. While prior work focuses  on extractive rationales (a subset of the input words), we investigate their less-studied counterpart: free-text natural language rationales. We demonstrate that \emph{pipelines}, models for faithful rationalization on information-extraction style tasks, do not work as well on ``reasoning’' tasks requiring free-text rationales. We turn to models that \textit{jointly} predict and rationalize, a class of widely used high-performance models for free-text rationalization. We investigate the extent to which the labels and rationales predicted by these models are associated, a necessary property of faithful explanation. Via two tests, \emph{robustness equivalence} and \emph{feature importance agreement}, we find that state-of-the-art T5-based joint models exhibit desirable properties for explaining commonsense question-answering and natural language inference, indicating their potential for producing faithful free-text rationales.\footnote{Our code is available at \url{https://github.com/allenai/label_rationale_association}.}
\end{abstract}

\section{Introduction}
\label{sec:introduction}

Interpretable NLP aims to better understand predictive models' internals for purposes such as debugging, validating safety before deployment, or revealing unintended biases and behavior \cite{molnar2019}. These objectives require faithful rationales---explanations of the model's behavior that are accurate representations of its decision process \cite{melis2018towards}. 

\begin{figure}[t]
  \centering
  \includegraphics[width=\columnwidth]{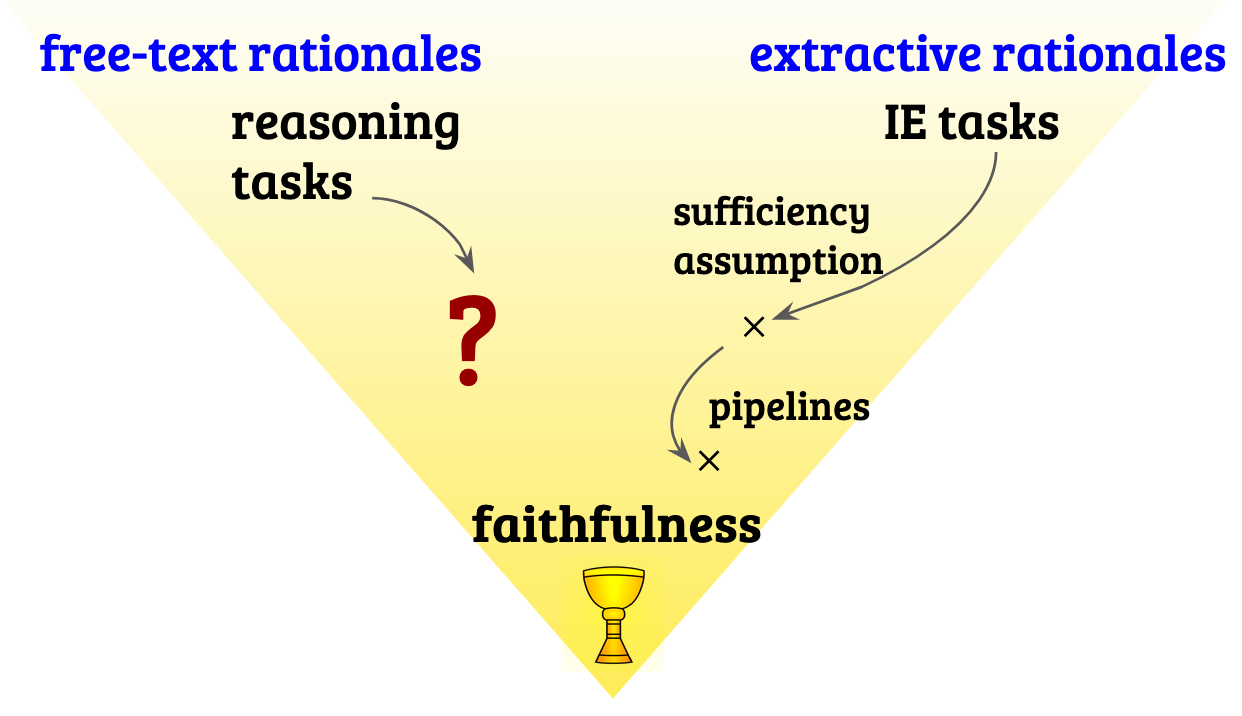}
  \caption{A categorization of interpretable NLP on an illustrative faithfulness spectrum.
  Two predominant forms of explanation exist that align with two predominant classes of NLP tasks. Unlike models for IE tasks, the desirable properties of interpretable models for reasoning tasks have not been explored. We investigate architectures and tests for explaining reasoning tasks.
  }
  \label{fig:motivation_fig}
\end{figure}
\begin{table*}
\resizebox{\textwidth}{!}{
\begin{tabular}{ll}
\toprule
\multirow{3}{2cm}{Commonsense QA (CoS-E)} & \textbf{Question:} While eating a \colorbox{yellow!30}{hamburger with friends}, what are people trying to do? \\
& \textbf{Answer choices:} \underline{have fun}, tasty, or indigestion \\
& \textbf{Natural language rationale:} Usually a hamburger with friends indicates a good time.                                             \\
\midrule
\multirow{3}{2cm}{Natural Language Inference (E-SNLI)} & \makecell[l]{\textbf{Premise:} A child in a yellow plastic safety swing is laughing as a dark-haired woman stands behind her.} \\
& \textbf{Hypothesis:} A young \colorbox{yellow!30}{mother} is playing with her \colorbox{yellow!30}{daughter} in a swing. \\
& \textbf{Label choices:} \underline{neutral}, entailment, or contradiction \\
& \textbf{Natural language rationale:} Child does not imply daughter and woman does not imply mother.\\
\bottomrule     
\end{tabular}}
\caption{Examples from the CoS-E v1.0 and E-SNLI datasets (\sect{sec:tasks_datasets}). Extractive rationales annotated by humans are highlighted, while human-written free-text rationales are presented underneath the answer/label choices. These examples illustrate that the extractive rationales fail to adequately explain the correct (underlined) label.}
\label{tab:datasets_all}
\end{table*}

One way towards faithfulness is to introduce architectural modifications or constraints that produce rationales with desirable properties \cite[\emph{inter alia}]{andreas2016neural, schwartz-etal-2018-bridging, jiang-etal-2019-explore}. For example, pipeline models (\autoref{fig:pipeline}) were designed for information extraction (IE) tasks for which a rationale can be extracted as a subset of the input and is \emph{sufficient} to make a prediction on its own, without the rest of the input \cite{lei-etal-2016-rationalizing}. Such models approach faithfulness by construction \cite{jain-etal-2020-learning}.

There is a growing interest in tasks that require world and commonsense ``knowledge'' and ``reasoning'', such as commonsense question-answering \cite[\cqa;][]{talmor-etal-2019-commonsenseqa} and natural language inference \cite[SNLI;][]{bowman-etal-2015-large}. Here, extractive rationales necessarily fall short---rationales must instead take the form of free-text natural language to fill in the reasoning or knowledge gap \cite{camburu2018snli, rajani-etal-2019-explain}.\footnote{We use ``free-text'' and ``natural language'' rationales interchangeably. We additionally use the term ``rationale'' to also mean ``explanation''; for a more detailed discussion of terminology see \citet{jacovi2020aligning, wiegreffe2021teach}.} In \autoref{tab:datasets_all}, for example, the highlighted extractive rationale of the first problem instance lacks at least one reasoning step to adequately justify the answer; the natural language rationale (which is not extractive) fills in the gap. 

We study two distinct model classes: \emph{self-rationalizing} models, which are fully differentiable and jointly predict the task output with the rationale; and \emph{pipelines}, which rationalize first and then predict task output with a separate model. We first show that, for CommonsenseQA and SNLI, a self-rationalizing model provides rationales that better indicate the correct label than a pipeline (\sect{sec:rationale_quality}). Next, we show that sufficiency is not universally applicable:  a natural language rationale  on its own does not generally provide enough information to arrive at the correct answer (\sect{sec:sufficiency}). These findings suggest that a faithful-by-construction pipeline is not an ideal approach for reasoning tasks, leading us to ask: is there is a way to achieve faithful free-text rationalization with self-rationalizing models? 

We note that there is currently no way to assess the relationship between a prediction and a free-text rationale within the same fully differentiable model. \citet{jacovi-goldberg-2020-towards} argue for the development of evaluations that measure the \emph{extent} and \emph{likelihood} that a rationale is faithful in practice (illustrated in \autoref{fig:motivation_fig}). To do so, we propose two measurements to initiate testing the extent to which predicted labels and explanations are associated within the model that produces them.

The first experiment, \emph{robustness equivalence} (\sect{sec:robustness}), analyzes whether a predicted label and generated rationale are similarly robust to noise. The second, \emph{feature importance agreement} (\sect{sec:gradients}), analyzes whether the gradient-attributions of the input with respect to the predicted label are similar to those with respect to the predicted rationale. We show that a self-rationalizing finetuned variant of T5 \cite{raffel2019exploring,narang2020wt5} demonstrates good robustness equivalence and feature importance agreement on the datasets investigated. This result motivates future work on more measurements for testing label-rationale association.

\section{Tasks, Datasets, and Models}\label{section2}

Before we turn to our analyses we introduce datasets and models used for our experiments.

\begin{table*}
    \centering
    \resizebox{\textwidth}{!}{
    \begin{tabular}{lccccccccccccc}
    \toprule
        \multirow{2}{*}{\textbf{Source}} & \multirow{2}{*}{\textbf{CQA}} & \multirow{2}{*}{\textbf{SNLI}} & \multirow{2}{*}{\textbf{SST}} & \multirow{2}{*}{\textbf{AgNews}} & \textbf{Evidence} & \textbf{Movie} & \multirow{2}{*}{\textbf{MultiRC}} & \multirow{2}{*}{\textbf{LGD}} & \multirow{2}{*}{\textbf{20 News}} & \textbf{Amazon} & \textbf{Beer} & \multirow{2}{*}{\textbf{BoolQ}} & \multirow{2}{*}{\textbf{FEVER}} \\
        & & & & & \textbf{Inference} & \textbf{Reviews} & & & & \textbf{Reviews} & \textbf{Reviews} \\
        \midrule
        \textbf{True Pipelines (no gradient flow)}  &  &  & & & & & & & & & & & \\
        \citet{camburu2018snli} &  \cellcolor{black!5}& E + NL & \cellcolor{black!5} &\cellcolor{black!5} &\cellcolor{black!5} & \cellcolor{black!5}& \cellcolor{black!5}& \cellcolor{black!5}& \cellcolor{black!5}&\cellcolor{black!5} &\cellcolor{black!5} &\cellcolor{black!5} &\cellcolor{black!5} \\
        \citet{kumar-talukdar-2020-nile} & \cellcolor{black!5} & NL & \cellcolor{black!5} &\cellcolor{black!5} &\cellcolor{black!5} & \cellcolor{black!5}& \cellcolor{black!5}& \cellcolor{black!5}& \cellcolor{black!5}&\cellcolor{black!5} &\cellcolor{black!5} &\cellcolor{black!5} &\cellcolor{black!5} \\
        \citet{rajani-etal-2019-explain} & E + NL & \cellcolor{black!5}& \cellcolor{black!5} &\cellcolor{black!5} &\cellcolor{black!5} & \cellcolor{black!5}& \cellcolor{black!5}& \cellcolor{black!5}& \cellcolor{black!5}&\cellcolor{black!5} &\cellcolor{black!5} &\cellcolor{black!5} &\cellcolor{black!5} \\
        \citet{jain-etal-2020-learning} & \cellcolor{black!5}&\cellcolor{black!5}& E & E & E & E & E & \cellcolor{black!5}& \cellcolor{black!5}& \cellcolor{black!5}& \cellcolor{black!5}& \cellcolor{black!5}& \cellcolor{black!5}\\
        \citet{jacovi2020aligning} & \cellcolor{black!5}& \cellcolor{black!5}& E & E & E & E & E & E & E & E & E & \cellcolor{black!5}&\cellcolor{black!5}\\
        \citet{deyoung-etal-2020-eraser} & E & E & \cellcolor{black!5}& \cellcolor{black!5}& E & E & E & \cellcolor{black!5}& \cellcolor{black!5}& \cellcolor{black!5}& \cellcolor{black!5}& E & E \\
        \citet{lehman-etal-2019-inferring} & \cellcolor{black!5}& \cellcolor{black!5}&\cellcolor{black!5}&\cellcolor{black!5}& E &\cellcolor{black!5}& \cellcolor{black!5}& \cellcolor{black!5}& \cellcolor{black!5}&\cellcolor{black!5}&\cellcolor{black!5}&\cellcolor{black!5}&\cellcolor{black!5}\\ \midrule
        \textbf{Discrete Optimization Variants}  &  &  & & & & & & & & & & & \\
        \citet{lei-etal-2016-rationalizing} & \cellcolor{black!5}& \cellcolor{black!5}& \cellcolor{black!5}& \cellcolor{black!5}& \cellcolor{black!5}& \cellcolor{black!5}& \cellcolor{black!5}& \cellcolor{black!5}& \cellcolor{black!5}& \cellcolor{black!5}& E &\cellcolor{black!5}&\cellcolor{black!5}\\
        \citet{bastings-etal-2019-interpretable} & \cellcolor{black!5}& E & E &\cellcolor{black!5}&\cellcolor{black!5}&\cellcolor{black!5}&\cellcolor{black!5}&\cellcolor{black!5}&\cellcolor{black!5}&\cellcolor{black!5}& E & \cellcolor{black!5}&\cellcolor{black!5}\\
        \citet{latcinnik2020explaining} & NL &  \cellcolor{black!5}& \cellcolor{black!5}& \cellcolor{black!5}& \cellcolor{black!5}& \cellcolor{black!5}& \cellcolor{black!5}& \cellcolor{black!5}& \cellcolor{black!5}& \cellcolor{black!5}& \cellcolor{black!5}& \cellcolor{black!5}& \cellcolor{black!5}\\
        \citet{paranjape-etal-2020-information} &\cellcolor{black!5} & \cellcolor{black!5}&\cellcolor{black!5}& \cellcolor{black!5}& E & E & E & \cellcolor{black!5}& \cellcolor{black!5}& \cellcolor{black!5}& E & E & E \\
    \bottomrule
    \end{tabular}
    }
\caption{An overview of text-only datasets and rationale types (E for extractive, NL for natural language rationales) used in prior work on pipeline architectures. We focus on the two tasks we believe require a more complex notion of ``reasoning'' to solve: CommonsenseQA (CQA) and NLI. Unlike the other tasks in the table, prior work for rationalizing these two tasks lacks consensus on (1) the type of rationales best-suited, and (2) the form of the model for these tasks. We argue for natural language rationales, and demonstrate that pipeline models are poorly-suited for CQA and SNLI given this choice. Dataset citations: Appendix \ref{sec:appendixcite}.}
    \label{table:datasets}
\end{table*}

\paragraph{Tasks and Datasets}
\label{sec:tasks_datasets}
We explore two large-scale datasets for textual reasoning tasks that contain human-written natural language rationales: E-SNLI \cite{camburu2018snli}, 
an extension of SNLI \cite{bowman-etal-2015-large}; and CoS-E \cite{rajani-etal-2019-explain}, an extension of \cqa \cite{talmor-etal-2019-commonsenseqa} (both in English). For the former, the task is to infer whether a given hypothesis sentence entails, contradicts, or is neutral towards a premise sentence. For the latter, the task is to select the correct answer from 3 (v1.0) or 5 (v1.11) answer choices for a question. We use both versions of CoS-E in our experiments (see Appendix \ref{sec:appendix_dataset}).  \autoref{tab:datasets_all} contains examples and 
\autoref{table:stats} (Appendix \ref{sec:appendix_dataset}) data statistics.\footnote{CoS-E does not contain test set rationale annotations, so we report performance values on the validation set.}

\begin{figure*}[t]
  \centering
  \includegraphics[width=\textwidth]{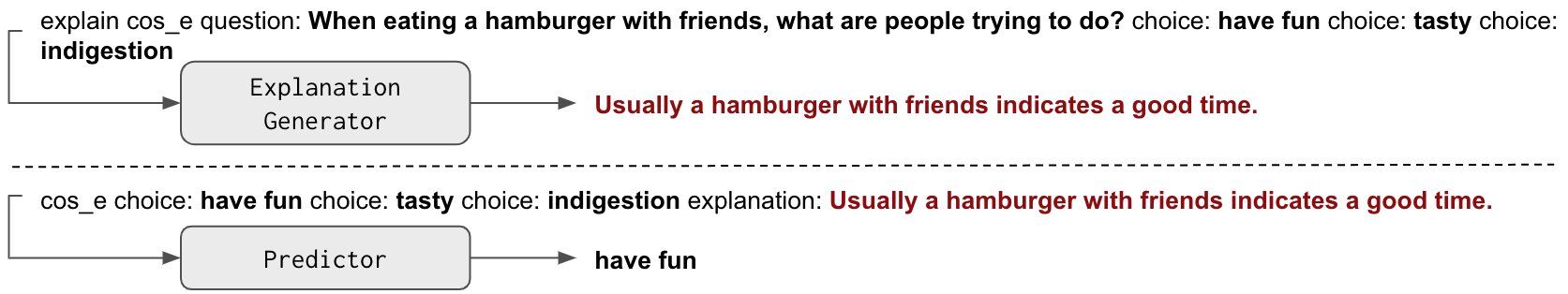}
  \caption{An illustration of a pipeline model (composed of \itor and \rtoo; \sect{sec:models}) for CoS-E v1.0 with a human-written rationale. The dotted line indicates two separate models with no gradient flow.
  } 
  \label{fig:pipeline}
\end{figure*}

\paragraph{T5 Models}
\label{sec:models}
All of the models in this work are based on T5, though our methods can in principle be applied to any architecture. The base version of T5 is a 220M-parameter transformer encoder-decoder \cite{vaswani2017attention}. To carry out supervised finetuning, T5 is trained by maximizing the conditional likelihood of the correct text output (from annotated data), given the text input.

We finetune five T5-Base models for each dataset, supervising with ground-truth labels and rationales (further details in Appendix \ref{sec:appendix_t5_details}-\ref{sec:hparams}):

 \begin{compactitem}
     \item \itor, which maps task inputs
 to rationales, without ever being exposed to task outputs.
     \item \rtoo, which maps rationales to task outputs.  The only input elements  this model is exposed to are
     answer choices (for CoS-E).
     \item \itoor, which maps inputs to outputs and rationales.  
     \item \irtoo, which maps pairs of inputs and rationales to outputs. 
    \item I$\rightarrow$O\xspace, which maps inputs to outputs. 
 \end{compactitem}

We provide input-output formatting in~\autoref{tab:t5_data_format} (Appendix \ref{sec:appendix_t5_details}). Using these building blocks, we can instantiate two important approaches.

\paragraph{Pipeline Model (\itor;\rtoo)}  
This architecture composes \itor with \rtoo, each of which is trained entirely separately, for a total of 440M parameters.
It is illustrated in
\autoref{fig:pipeline} and is faithful-by-construction (with caveats; see \citealp{jacovi2020aligning}).
The vast majority of prior work using pipelines has focused only on extractive rationales (see~\autoref{table:datasets}).

\begin{figure*}
  \centering
  \includegraphics[width=\linewidth]{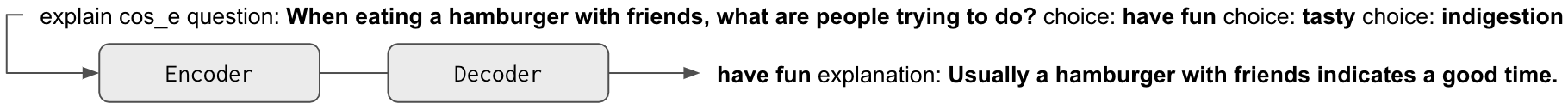}
  \caption{An example of a joint architecture (\itoor; \sect{sec:models}) for CoS-E v1.0 with a human-written rationale. Trained on both task signal and human rationales, these models are effective at generating fluent rationales while retaining good task performance (\autoref{table:baselines}-\ref{table:pipelines} in \autoref{sec:appendix_results}).
  }
  \label{fig:joint}
\end{figure*}

\paragraph{Self-Rationalizing Model (\itoor)} \label{sec:self_rationalizing}  A joint, self-rationalizing model \cite{melis2018towards}, illustrated in \autoref{fig:joint}, predicts both a label and rationale. This is the most common approach to free-text rationalization \cite{hendricks2016generating, kim2018textual, hancock-etal-2018-training, camburu2018snli, ehsan2018rationalization, liu-etal-2019-towards-explainable, wu-mooney-2019-faithful, narang2020wt5, Do2020eSNLIVE20CV, tang-etal-2020-exploring}, but little is understood about model internals. \itoor models are desirable for their ease-of-use, task-effectiveness, parameter efficiency, and their ability to generate fluent and plausible rationales. 
We expect models of this kind to play an important role in continuing research on explainable AI for these reasons.

We use the \itoor variant of T5 \cite{narang2020wt5}. Because only one instance of T5 is used to instantiate it, the total number of parameters is half that of the pipeline. We replicate two prior findings (Tables \ref{table:baselines}--\ref{table:pipelines} in \autoref{sec:appendix_results}): the T5 pipeline does not perform as well as the self-rationalizing model (despite having double the parameters), and T5-Base outperforms pretrained models used in prior work.

\paragraph{Evaluation} \label{sec:evaluation}

We do not report BLEU scores \cite{papineni-etal-2002-bleu}, because BLEU and related metrics do not measure plausibility \cite{camburu2018snli, kayser2021vil, clinciu-etal-2021-study} or faithfulness \cite{jacovi-goldberg-2020-towards}. In addition to low correlation with human scores, there can be many valid rationales for a given instance \cite{miller2019explanation}; metrics that compare generated rationales to a single ground-truth do not address this and are thus a poor measure of quality.

Human simulatability \cite{doshi2017towards} has a rich history in machine learning interpretability as a reliable measure of rationale quality from the lens of \emph{utility to an end-user} \cite[\emph{i.a.}]{kim2016examples, chandrasekaran-etal-2018-explanations, hase-bansal-2020-evaluating, yeung2020sequential, poursabzi2018manipulating, rajagopal2021selfexplain}. Rather than computing word-level overlap with a ground-truth explanation, simulatability measures the additional predictive ability towards the predicted label a rationale provides over the input, computed as the difference between task performance when a rationale is given as input vs.\ when it is not (IR$\rightarrow$\^{O}\xspace minus I$\rightarrow$\^{O}\xspace).\footnote{The predicted label is from the same system that produced the predicted rationale.}
Historically, humans have served as the predictors, but recent work has shown that the computation of simulatability can be automated using trained models. \citet{hase-etal-2020-leakage} demonstrate that automated metrics for simulatability have moderate to high correlation with human scores in both an expert and a crowdsourced setting. In our experiments, model predictions are often unable to be simulated because they degenerate under high values of noise (\sect{sec:metrics}, \ref{ssec:convergence}). We thus use a variant of this metric that relies on predicting the \emph{gold} labels as our measure of rationale quality: IR$\rightarrow$O\xspace minus I$\rightarrow$O\xspace.\footnote{Given the large scale of our analysis (>250K instances evaluated), an automated metric provides coverage, reproducibility and consistency not achievable with human annotation. An author of this paper annotated 60 instances from both E-SNLI and CoS-E v1.11, and found 82.5\% agreement between their rationale quality score and the automated metric.} We discuss the effects of this difference in Appendix \ref{ssec:caveats}.

\section{Shortcomings of Free-Text Pipelines}
\label{sec:abstraction}

We first analyze ``faithful-by-construction'' 
pipeline models (\itor;\rtoo) for free-text rationalization with respect to two properties: quality of generated rationales (\sect{sec:rationale_quality}) and appropriateness of the sufficiency assumption (\sect{sec:sufficiency}).

\subsection{Joint Model Rationales are More Indicative of Labels}
\label{sec:rationale_quality}

\begin{table}
    \centering
    \resizebox{0.9\linewidth}{!}{
    \begin{tabular}{|l|l|l|l|}
    \toprule
        \textbf{Source of Rationales} & \textbf{R$^{*}$} & \textbf{\itor} & \textbf{\itoor}\\
        \midrule
        E-SNLI & 97.67 & 89.11 & 90.52 \\
        CoS-E v1.0 & 84.84 & 53.47 & 62.00 \\
        CoS-E v1.11 & 68.63 & 45.45 & 53.15 \\
        \bottomrule
    \end{tabular}}
    \caption{Accuracy of the trained \rtoo model evaluated on ground-truth natural language rationales ($\text{R}^{*}$) and rationales generated from two model architectures: \itoor and \itor (see \sect{sec:models} for model descriptions).
    }
    \label{table:generated-rationales}
\end{table}
\begin{table}
    \centering
    \resizebox{0.9\linewidth}{!}{
    \begin{tabular}{|l|l|l|l|}
    \toprule
        \textbf{Source of Rationales} & \textbf{R$^{*}$} & \textbf{\itor} & \textbf{\itoor}\\
        \midrule
        E-SNLI & 7.77 & -1.63 & -0.86 \\
        CoS-E v1.0 & 21.26 & -12.11 & -6.21 \\
        CoS-E v1.11 & 19.09 & -12.77 & -6.06 \\
        \bottomrule
    \end{tabular}
    }
    \caption{Rationale quality scores (\sect{section2}; higher is better) of ground-truth rationales ($\text{R}^{*}$) and rationales generated from two model architectures: \itoor and \itor. These results demonstrate that rationales generated as a function of the input and the predicted label (\itoor) are higher quality than those generated as a function of the input alone (\itor) across datasets (\sect{sec:rationale_quality}).
    }
    \label{table:las}
\end{table}

Rationales should be a function of the input \emph{and} the predicted label. To demonstrate why this is the case, consider training an \itor model on a dataset with multiple annotation layers, e.g., OntoNotes, that contains word sense, predicate structure, and coreference \cite{pradhan2007ontonotes}. Without additional task-specific input, this model would produce the same rationale, regardless of the task being rationalized. Prior work has also critiqued \itor;\rtoo models because it is counter-intuitive to generate a rationale before deciding the label to explain \cite{kumar-talukdar-2020-nile, jacovi2020aligning}. Therefore, the \itor model will first need to implicitly predict a label. But can \itor infer the label well, when it is trained without label signal?

To address this question, we study whether \itoor rationales are better at predicting the gold labels than \itor rationales. We train a \rtoo model on ground-truth rationales (R$^{*}$), and evaluate on the following inputs:
\begin{compactitem}
    \item test set ground-truth R$^{*}$ rationales,
    \item test set rationales generated by \itoor, and 
    \item test set rationales generated by \itor.
\end{compactitem}

In Table 2, we show that \itoor rationales recover 8--9\% more ground-truth (R$^*$) performance than \rtoo rationales on both versions of CoS-E, and 1\% on E-SNLI.
A smaller improvement for E-SNLI could be explained by the fact that E-SNLI has substantially more training examples for each label than CoS-E, which helps a pipeline model learn features predictive of each label.\footnote{E-SNLI has 549,357 training examples and only 3 labels. In contrast, the number of answer options across all instances in CoS-E v1.0 (v1.11) is 6,387 (12,992), but the training set size is 7,610 (9,741), i.e., $\sim$56 (72) times smaller than E-SNLI.}
We additionally demonstrate \itoor rationales are higher quality than \rtoo's, as measured by our rationale quality metric (\autoref{table:las}).
The fact that the pipeline's strong performance does not generalize to a complex prediction task such as CoS-E empirically demonstrates that training on label signal O is important to generate good-quality rationales and avoid cascading errors.

\subsection{Sufficiency is not  Universally Valid}

\label{sec:sufficiency}

\begin{table}
    \centering
    \resizebox{\columnwidth}{!}{
    \begin{tabular}{|l|r|r|r|}
    \toprule
        \textbf{Model} & \textbf{\rtoo with R$^*$} & \textbf{\irtoo with R$^*$} & \textbf{\emph{$\Delta$}}\\
        \midrule
        E-SNLI & 97.67 & 98.72 & +1.05 \\
        CoS-E v1.0 & 84.84 & 90.42 & +5.58 \\
        CoS-E v1.11 & 68.63 & 80.84 & +12.21 \\
        \bottomrule
    \end{tabular}}
    \caption{A comparison of the \irtoo and \rtoo models (\sect{sec:models}) evaluated with ground-truth natural language rationales ($\text{R}^{*}$). In some cases accuracy improves substantially with the addition of the input, indicating that rationales are not always sufficient and pipelines are not always effective.}
    \label{table:complementary-vs-standalone}
\end{table}

``Faithful-by-construction'' pipelines rely on the \emph{sufficiency} assumption: the selected rationale must be sufficient to make the prediction without the remaining input. This assumption is suitable for IE tasks for which a subset of the input tokens is predictive of the label. Indeed, humans can serve as  \rtoo models on certain IE tasks and make accurate predictions, validating that rationales are sufficient for these tasks \cite{jain-etal-2020-learning}.

To illustrate why sufficiency might not be justified for reasoning tasks, consider the  example in Figure \ref{fig:pipeline}. The task of the \rtoo model is to select between the answer choices ``have fun'', ``tasty'', and ``indigestion'' given the rationale ``Usually a hamburger with friends indicates a good time''. The rationale is designed to complement the input question, but the \rtoo model does not see the question, changing the fundamental nature of the task it is solving. We thus wonder: does task obfuscation hurt pipelines' ability to perform the task?

We report the accuracy difference between a \rtoo model and a model that receives both the input and rationale (\irtoo), both trained on R$^{*}$. We evaluate on test set R$^{*}$.\footnote{Evaluating on R$^{*}$ instead of generated rationales serves as an upper-bound on pipeline performance, removing the confounding factor that \itor rationales can be poor (\sect{sec:rationale_quality}).} In \autoref{table:complementary-vs-standalone}, the \irtoo models on CoS-E have a 5--12\% increase in accuracy over  \rtoo, indicating that the rationales are not sufficient. The difference is much smaller for E-SNLI (1\%), likely due to the fact that E-SNLI was collected by instructing annotators to provide self-contained rationales.
However, using dataset-collection to explicitly collect sufficient rationales does not address the unnaturalness of such a task formulation \cite{wiegreffe2021teach}.\footnote{\citet{camburu2018snli} give an example: the rationale ``A woman is not a person'' could predict either a contradiction or entailment label depending on the input.} \autoref{table:complementary-vs-standalone} indicates that (especially) in the case of CoS-E, sufficiency is not a valid assumption, and the use of \itor;\rtoo models is sub-optimal in these cases. 

So far, we have highlighted shortcomings of pipelines for reasoning tasks:
\begin{compactitem}
    \item cascading errors caused by low-quality rationales that are not indicative of labels (\sect{sec:rationale_quality}),
    \item missing information due to rationales not being sufficient (\sect{sec:sufficiency}),
    \item double the number of parameters and more manual labor needed to reach comparable performance to an end-to-end (I$\rightarrow$O) model; still often performing worse (\sect{sec:tasks_datasets}). 
\end{compactitem}

We next turn our focus to self-rationalizing (\itoor) models currently in widespread use, which in contrast to pipelines are high-performing, easy to implement via a multi-task loss, and more parameter-efficient (\sect{sec:tasks_datasets}).

\section{Analyzing Necessary Properties of Joint Models}
\label{sec:metrics}

Despite their popularity and widespread use, the extent to which  self-rationalizing models exhibit faithful rationalization has not been studied. To illustrate this point, we reference \citet{narang2020wt5}:
\begin{quotation}
    \noindent \small \dots Much like humans, our approach does not guarantee that the produced explanation actually explains the specific reasons why a model generated its prediction. In other words, the model could potentially just make up a reasonable-sounding explanation instead  of providing a truly accurate description of its causal decision-making process.
\end{quotation}
It is not infeasible that a large, overparameterized model trained on both gold-rationale emulation and a labelling task can learn to do both equally well, without having to rely on shared information in its parameters. Therefore, rationales from \itoor models cannot be treated as faithful explanations without further investigation.

At minimum, rationales must be implicitly or explicitly tied to the model's prediction. 
We present two metrics to analyze the association between the mechanisms that produce labels and rationales in a multi-task, \itoor model:
\emph{robustness equivalence} (\sect{sec:robustness}) and \emph{feature importance agreement} (\sect{sec:gradients}). These experiments serve as a necessary sanity check for the reliability of \itoor models' explanations.

\subsection{Robustness Equivalence}
\label{sec:robustness}

\begin{table*}
    \centering
    \resizebox{\textwidth}{!}{
    \begin{tabular}{l|l}
    \toprule
        \textbf{$\sigma^2$} & \textbf{Predicted Output}\\ 
        \midrule
        0 & have fun \textbf{explanation:} having fun is the only thing that people are trying to do.\\
        5 & have fun \textbf{explanation:} having fun is the only thing that people are trying to do.\\
        10 & have fun \textbf{explanation:} eating a hamburger with friends is fun.\\
        15 & have fun \textbf{explanation:} having fun is the only thing people are trying to do while eating a hamburger with friends.\\ 
        20 & <extra\_id\_0> a hamburger with friends<extra\_id\_1> are people trying to do? \textbf{explanation:} a hamburger is a hamburger\dots\\
        25 & "" is the only thing that is """""""""""""""""""""""""""""""""""""""""""""""""""""""""""""""""""""""""""""""""""""""""""\\
        30 & a indigestion: a indigestion: a indigestion: a indigestion: a indigestion: a indigestion: a indigestion: a indigestion: a indigestion:\dots\\ 
        35 & a ''''''''''''''''''''''''''''''''''''''''''''''''''''''''''''''''''"""""""""""""""""""""""""""""" \\
        \bottomrule
    \end{tabular}
    }
    \caption{Output of the \itoor model for the running CoS-E v1.0 example under differing noise levels.
    While the rationale changes from variance 0-15, it is still valid for the given (correct) predicted label. At a variance of 20 and beyond, the model fails to generate both the correct label and a valid rationale. The model's predictions for this instance exhibit robustness equivalence.}  
    \label{table:noise_examples}
\end{table*}

\begin{figure}[t]
\centering
\begin{subfigure}[t]{0.48\textwidth}
\includegraphics[width=\linewidth]{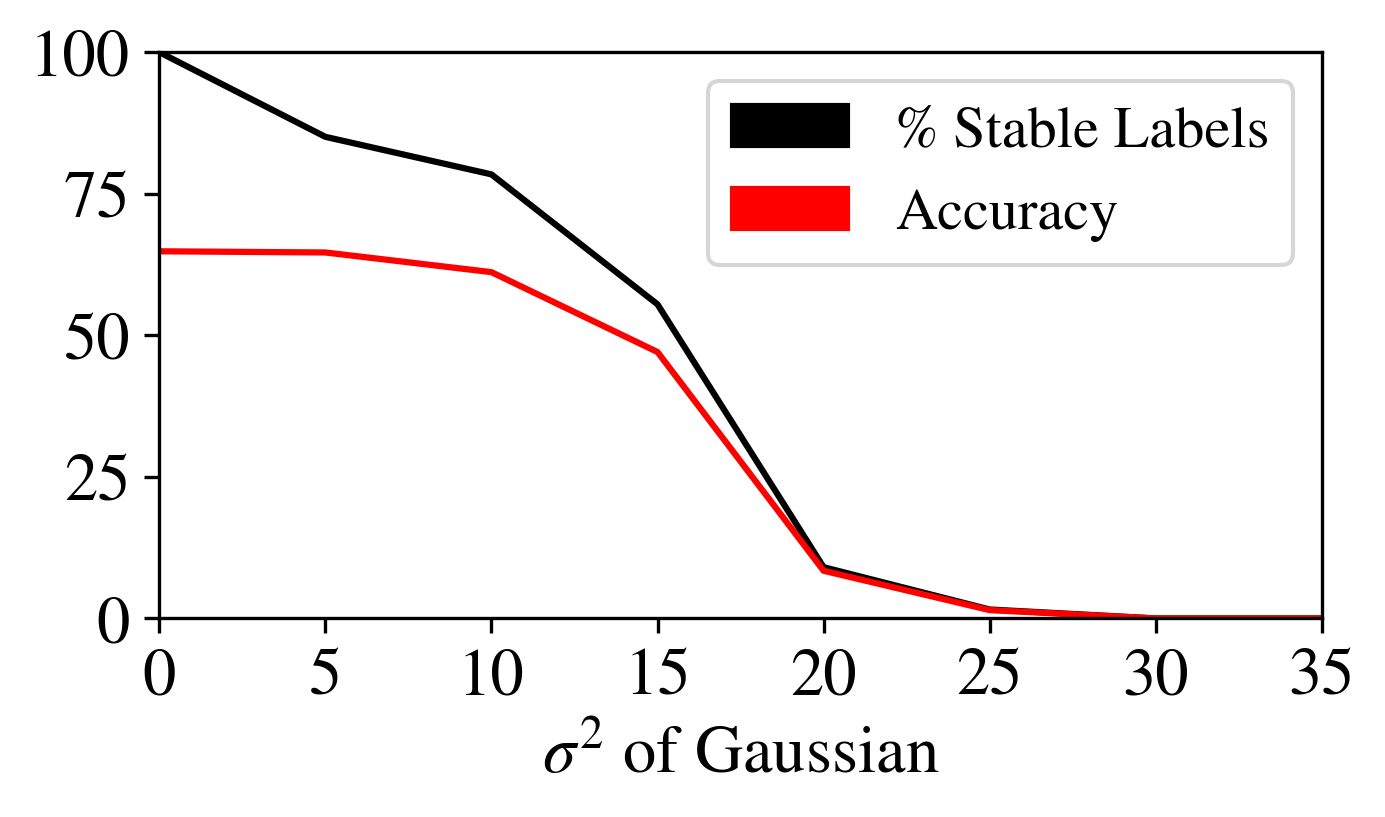}
\end{subfigure}
\caption{Results of the label portion of the robustness equivalence test for CoS-E v1.0.}
\label{fig:robustness} 
\end{figure}

\begin{figure}[t]
\centering
\begin{subfigure}[t]{0.48\textwidth}
\includegraphics[width=\linewidth]{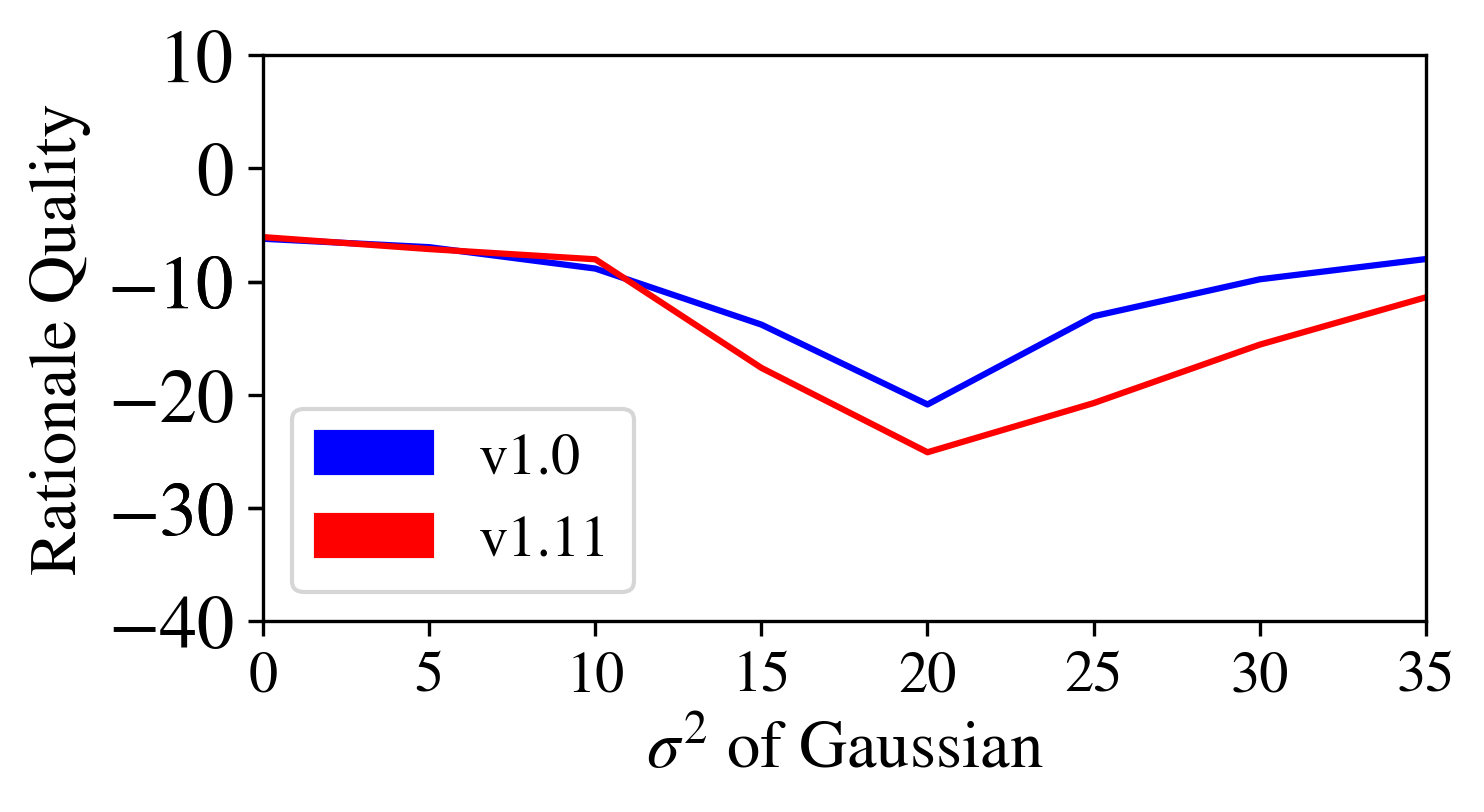}
\end{subfigure}
\caption{Results of the rationale portion of the robustness equivalence test for both CoS-E datasets.
}
\label{fig:robustness_label_flip}
\end{figure}

We aim to analyze whether predicted labels and rationales are similarly or dissimilarly robust to noise applied to the input. The former indicates predicted labels and rationales are strongly associated, while the latter indicates the opposite. Given some amount of noise, there are four possible cases for a model's output:
\(
     \{l_{\text{stable}},  l_{\text{unstable}}\} \times \{ r_{\text{stable}},  r_{\text{unstable}}\}
\),
where \(l\) is a label and \(r\) is a rationale.

The case where $l$ and $r$ are both stable or both unstable indicates that both tasks are similarly affected by noise. The case where $l$ is unstable but $r$ is stable (or vice versa) is a failure case---if only one output is stable, we conclude the two generation mechanisms cannot be strongly associated within the model.

\begin{figure}[t]
\centering
\begin{subfigure}[t]{0.8\columnwidth}
\includegraphics[width=0.8\linewidth]{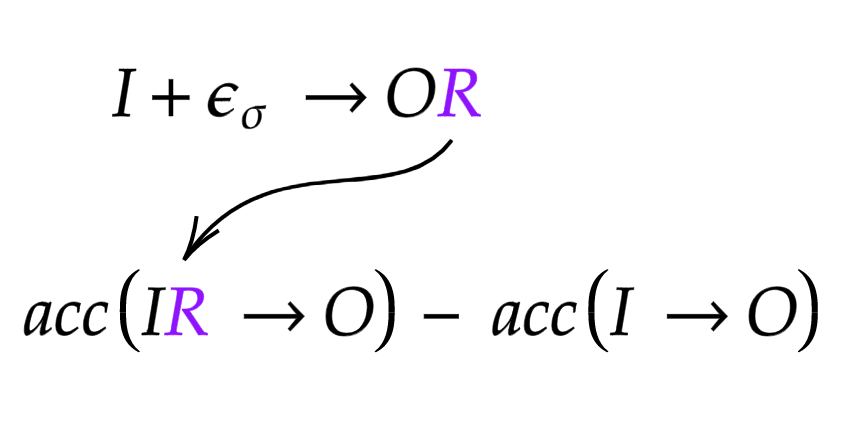}\caption{In \sect{sec:robustness}. $\epsilon_{\sigma}$ is the noise.}\label{fig:4_1_simulatability}
\includegraphics[width=0.8\linewidth]{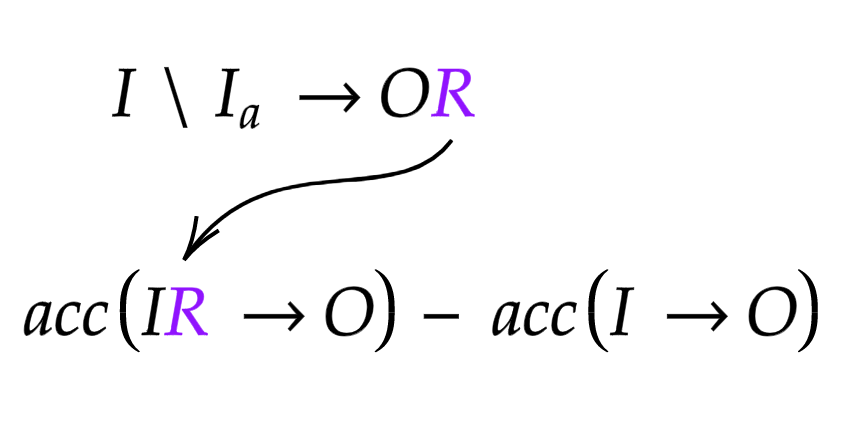}\caption{In \sect{sec:gradients}. $I_a$ are input tokens selected by an attribution method.}\label{fig:4_2_simulatability}
\end{subfigure}
\caption{An illustration of how rationale quality is calculated in \sect{sec:metrics}.}
\label{fig:simulatability}
\end{figure}

\paragraph{Method} Following related work \cite[]{pmlr-v97-wang19f, lakshmi-narayan-etal-2019-exploration, liu-etal-2019-zero}, we add zero-mean Gaussian noise $\mathcal{N}(0,\sigma^2)$ to each input embedding in the \itoor encoder at inference time. We measure changes in label prediction as the number of predicted test set labels that flip, i.e.,  change from their original prediction to something else, alongside changes in accuracy of the \itoor model. We measure changes in rationale quality using our rationale quality metric (\sect{sec:evaluation}), with details of the metric calculation illustrated in Figure \ref{fig:4_1_simulatability}. We report metrics on rationales generated by \itoor under different levels of noise, controlled by $\sigma^2$.

An example of noisy outputs for the running CoS-E v1.0 example is presented in \autoref{table:noise_examples}.

\paragraph{Results} We present results on the effect of noise on labels in \autoref{fig:robustness} (E-SNLI and CoS-E v1.11 in~\autoref{fig:cose-v2-robustness} of~\autoref{sec:appendix_results}).
As expected, the accuracy of the \itoor model (red line) and the percent of labels in the \itoor model which have not flipped (black line) are almost identical for all three datasets.
We present results on the effect of noise on rationales in \autoref{fig:robustness_label_flip} for CoS-E (E-SNLI in~\autoref{fig:robustness_label_flip_appendix} of~\autoref{sec:appendix_results}).

By examining the regions of largest slope, we gain insights into model behavior. 
On the rationale quality measure, both versions of CoS-E's rationales reach a minimum contribution to task accuracy at $\sigma^2 = 20$ (\autoref{fig:robustness_label_flip}). We similarly observe the largest drop in task accuracy (\autoref{fig:robustness}) for CoS-E v1.0 between $\sigma^2=15$ to $\sigma^2=20$.\footnote{See Appendix \ref{ssec:convergence} for a note on why the \itoor models achieve worse-than-random accuracy at high values of $\sigma^2$.} Thus, at lower noise levels (0--15), the model exhibits both stable labels and rationales, and at higher levels (20+), both unstable, indicating robustness equivalence. Similar conclusions can be reached for E-SNLI and CoS-E v1.11; we conclude that the \itoor model demonstrates high label-rationale association for all 3 datasets as measured by robustness equivalence.

We expect that rationale quality in \autoref{fig:robustness_label_flip} does not monotonically decrease because as rationales continue to worsen in quality (see the example in  \autoref{table:noise_examples}), the \irtoo model may ignore them completely and more closely emulate the I$\rightarrow$O model. For example, in \autoref{table:noise_examples}, at $\sigma^2 = 30$, the rationale provides signal for an incorrect answer choice (``indigestion'') that does not exist at $\sigma^2 = 35$. Therefore, we consider rationales produced with $\sigma$ larger than the value at which the metric reaches the minimum as unstable (see further examples corroborating this in \autoref{sec:appendix_results}, Tables~\ref{table:cose_noise_1}--\ref{table:cose_noise_3}).

\subsection{Feature Importance Agreement} 
\label{sec:gradients}

If label prediction and rationale generation are associated, input tokens important for label prediction should be important for rationale generation and vice versa. We refer to this property as \textit{feature importance agreement}. To measure to what extent \itoor models exhibit this property, we use gradient-based attribution \cite{baehrens2010explain, simonyan2013deep} to identify tokens important for label prediction, and the \textbf{R}em\textbf{o}ve \textbf{a}nd \textbf{R}etrain (\textbf{ROAR}) occlusion method \cite{hooker2019benchmark} to analyze their impact on rationale generation (or vice versa).

\paragraph{Gradient Attribution}\label{ssec:computing} For a predicted class $p$, gradient attribution is a function of 
the gradient of the predicted class' logit $l_p$ with respect to an input token embedding $\mathbf{x}^{(i)} \in  \mathbb{R}^{d}$: 
\begin{equation}
    a(\mathbf{x}^{(i)}; l_p) = f(\nabla_{\mathbf{x}^{(i)}}{l_p}) \in \mathbb{R},\label{eq:token_attribution}
\end{equation}
where the function $f$ reduces the gradient to a scalar.
Choices for $f$ include $L_1$ or $L_2$ norm \cite{atanasova-etal-2020-diagnostic}, or an element-wise sum \cite{wallace-etal-2019-allennlp}.
Intuitively, the gradient measures how much an infinitesimally small change in the input changes the predicted class' logit, using a first-order Taylor series approximation of the logit function. 
Such methods have been extended to sequence-output models such as neural machine translation \cite{he-etal-2019-towards,ding-etal-2019-saliency, li-etal-2020-evaluating} by computing the sum of $m$ decoded logits $\{l_p^{(k)}\}_{k=1}^m$ with respect to the input: 
\begin{equation}
a(\mathbf{x}^{(i)}; \{l_p^{(k)}\}_k) = \textstyle\sum_{k=1}^m a(\mathbf{x}^{(i)}; l_p^{(k)}) \in \mathbb{R}. \label{eq:1}\\
\end{equation}
The attribution of a sequence of $n$ input token embeddings, $\mathbf{X} \in \mathbb{R}^{n \times d}$, is a vector $a(\mathbf{X}) = [a(\mathbf{x}^{(1)}),\hdots,a(\mathbf{x}^{(n)})] \in \mathbb{R}^n$, where $a(\mathbf{x}^{(i)})$ is shorthand for the value defined in ~\autoref{eq:1}.

By decomposing the term in \autoref{eq:1} into two parts, we obtain two attribution vectors over the input tokens; one for the predicted label logits $\mathcal{L}$, and one for the predicted rationale logits $\mathcal{R}$ in the decoded output: 

{\small
  \setlength{\abovedisplayskip}{6pt}
  \setlength{\belowdisplayskip}{\abovedisplayskip}
  \setlength{\abovedisplayshortskip}{0pt}
  \setlength{\belowdisplayshortskip}{3pt}
\begin{align*}
a(\mathbf{x}^{(i)}; \{l_p^{(k)}\}_k) &= \sum_{k \in \mathcal{L}} a(\mathbf{x}^{(i)}; l_p^{(k)}) + \sum_{k \in \mathcal{R}} a(\mathbf{x}^{(i)}; l_p^{(k)}),\\
a(\mathbf{X}) &= a(\mathbf{X})_\mathcal{L} + a(\mathbf{X})_\mathcal{R}.\numberthis\label{eq:3}
\end{align*}
}
\normalsize

\paragraph{Reliability of Gradient Attribution}
\label{sec:occlusion}
Before we measure feature importance agreement, it is critical to evaluate whether the gradient-attribution scores truly capture token importance, since these methods can be unreliable for certain datasets or architectures \cite{kindermans2019reliability}.
To validate that our attributions are reliable, we perform the ROAR test \cite{hooker2019benchmark}. Using attribution scores, we obtain the top-$k\%$ attributed tokens for every instance and occlude them following T5's pretraining procedure and mask tokens. We retrain a model on the occluded training set and evaluate on the occluded test set. We repeat this procedure for $k \in \{10\%,20\%,30\%\}$, and compare the drop in performance as $k$ increases to a baseline in which $k\%$ random tokens are dropped. To the extent that the occluded model fails to match the random model's performance, we can attribute such degradation to the removal of tokens that the original model finds informative.
A large drop in performance indicates that gradient attributions successfully identify important tokens in the input.

We first use this method to select an optimal gradient-attribution method and $f$ function (\autoref{fig:roar_testing_methods} in~\autoref{sec:appendix_results}). We find the $L_1$ norm of the embedding vector as $f$ to outperform the element-wise sum (which may suffer from dampened magnitudes). Unlike prior work in computer vision \cite{hooker2019benchmark}, we find raw gradients to perform comparably to the input*gradient variant \cite{shrikumar2017learning}. Thus, we compute attributions in subsequent experiments following Equation \ref{eq:token_attribution} with $f$ equal to the $L_1$ norm.

\begin{figure*}[t]
\centering
  \includegraphics[width=0.49\linewidth]{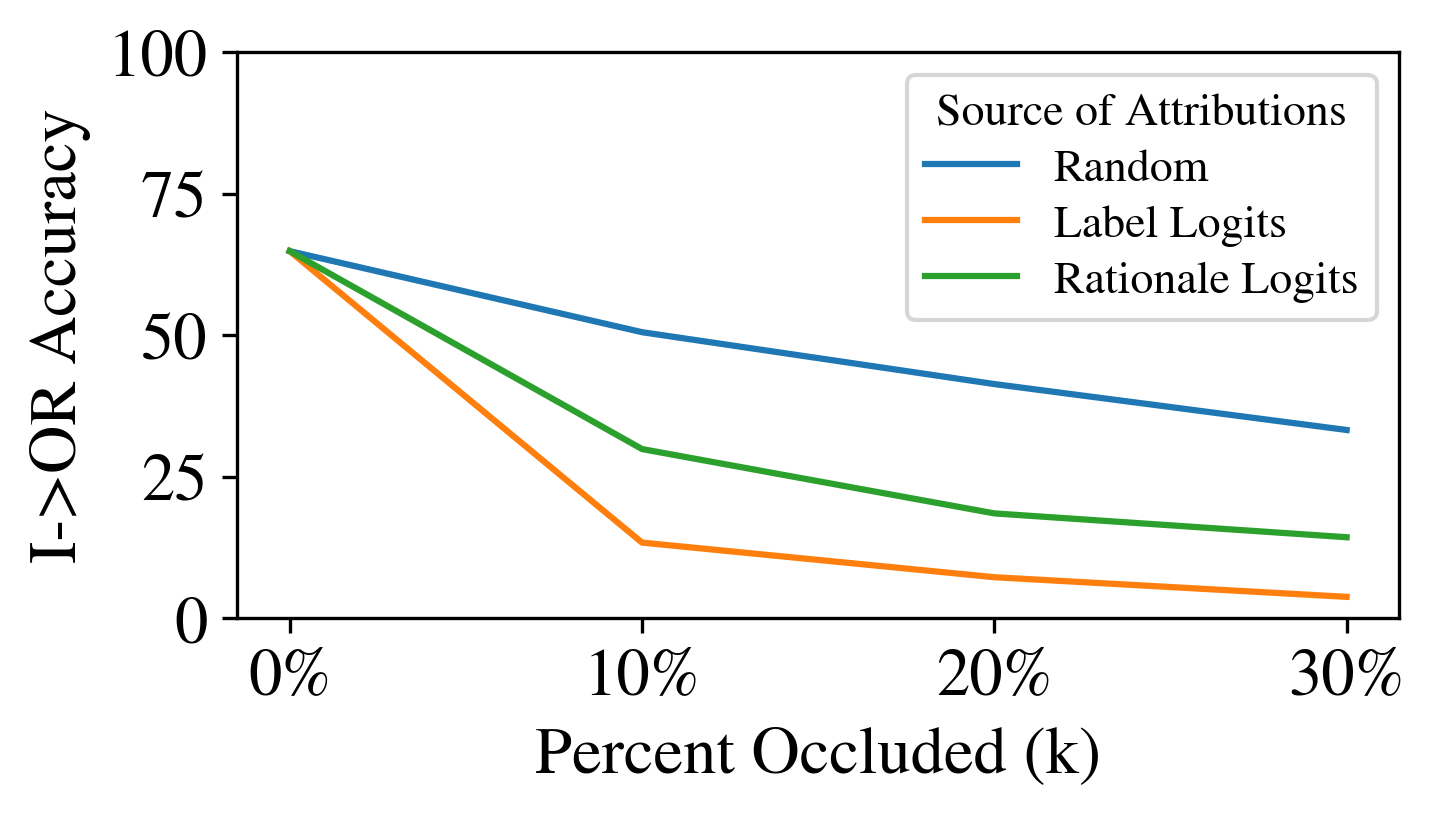}
  \includegraphics[width=0.49\linewidth]{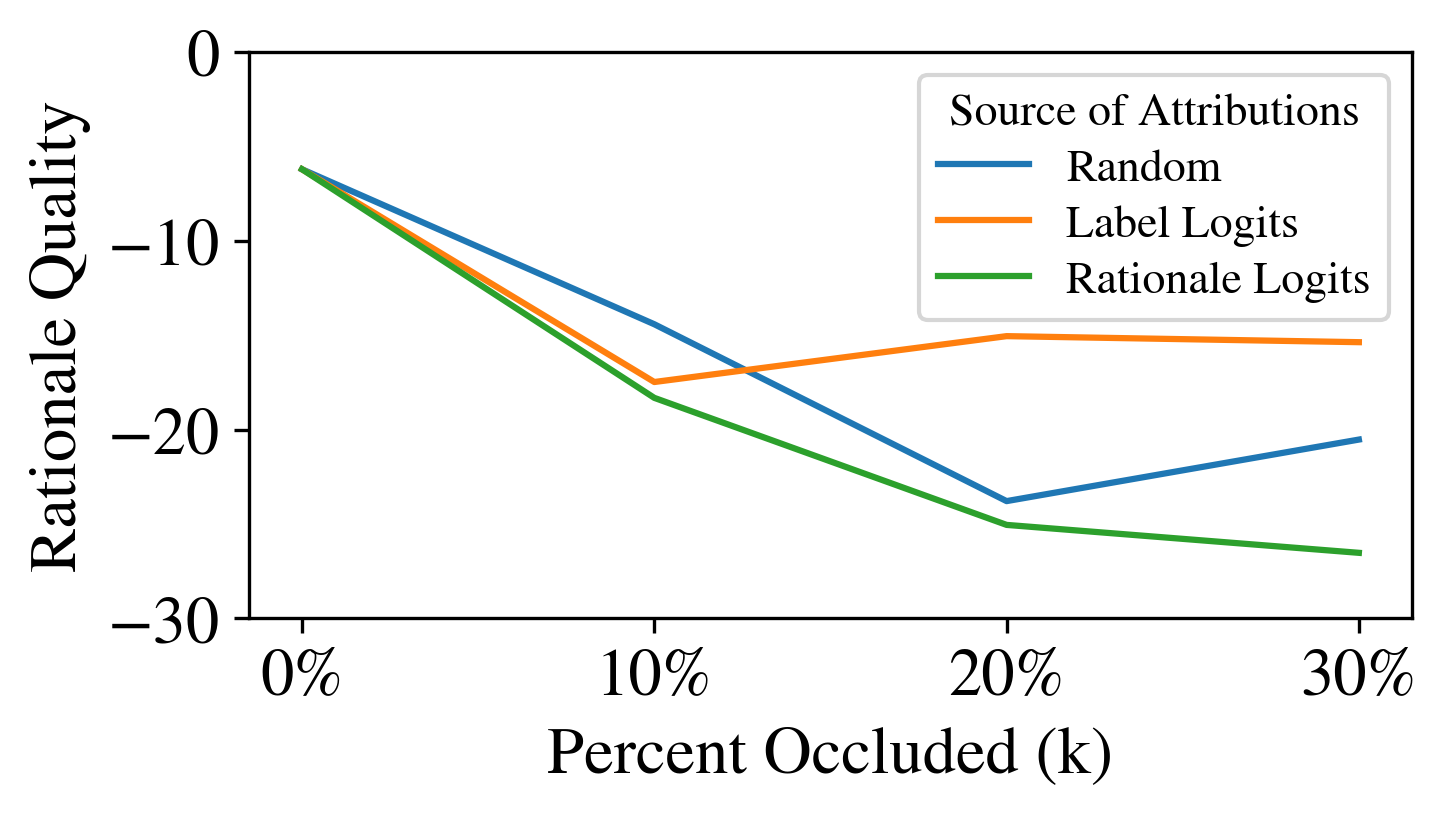}
\caption{Performance of \itoor models trained with the ROAR method 
on CoS-E v1.0. \textbf{Left:} Impact of occlusion by source of attribution on label accuracy. \textbf{Right:} Impact of occlusion by source of attribution on 
rationale quality.}
\label{fig:cose_v1_roar}
\end{figure*}

We validate that attributions from the label logits, $a(\mathbf{X})_\mathcal{L}$, degrade label accuracy when compared to random occlusion (orange vs.\ blue line in \autoref{fig:cose_v1_roar}, left). The two rationale quality lines (\autoref{fig:cose_v1_roar}, right) for CoS-E v1.0 have an inflection point. We illustrate how the metric is calculated in Figure \ref{fig:4_2_simulatability}. Similar to \sect{sec:robustness}, we expect this is due to rationales so noisy that \irtoo ignores them and behaves like I$\rightarrow$O. If an input attribution degrades rationale quality more than a random attribution, then the line corresponding to that attribution (for values of $k$ for which neither that attribution nor the random attribution have reached the inflection point) has to be below the ``random'' line. For values of $k$ for which both attributions have passed the inflection point, the ``random'' line should be below the attribution line, assuming that after this point, a noisier rationale is more similar is \irtoo to I$\rightarrow$O and hence the score is closer to 0. Both criteria hold for attributions from the rationale logits, $a(\mathbf{X})_\mathcal{R}$ (green vs.\ blue line in \autoref{fig:cose_v1_roar}, right) for CoS-E v1.0 and other datasets (see \autoref{fig:supp_roar_results} in~\autoref{sec:appendix_results}).
This reliability check confirms that gradient-attribution works well in our setting.

\paragraph{Agreement Method and Results}

To measure feature importance agreement---whether tokens important for label prediction are important for rationale generation (and vice versa)---we repeat the same experiment, but measure performance with respect to the \emph{other} output's metric. For attributions computed from label logits, $a(\mathbf{X})_\mathcal{L}$, we measure the effect of their occlusion on rationale quality using the rationale quality score. For attributions with respect to rationale logits, $a(\mathbf{X})_\mathcal{R}$, we measure the effect of their occlusion on label accuracy. If at least one of these values is notably different from random, we can conclude that the \itoor model displays feature-importance similarity in a given direction.

Results for CoS-E v1.0 are once again in \autoref{fig:cose_v1_roar} (and for other datasets in \autoref{fig:supp_roar_results} in~\autoref{sec:appendix_results}). In \autoref{fig:cose_v1_roar} (left), we find that removing top-$k\%$ tokens by $a(\mathbf{X})_\mathcal{R}$ magnitude degrades label performance compared to the baseline (green vs.\  blue line). Intuitively, this drop is less than token attributions by $a(\mathbf{X})_\mathcal{L}$ magnitude (orange line).
In \autoref{fig:cose_v1_roar} (right), we observe that removing top-$k\%$ tokens by $a(\mathbf{X})_\mathcal{L}$ consistently degrades rationale performance more than random according to the two criteria for comparing the rationale quality lines (orange  vs.\ blue line).
This also holds for E-SNLI and CoS-E v1.11 (\autoref{fig:supp_roar_results}). We conclude  that  the  \itoor  model  demonstrates label-rationale  association as measured by feature importance agreement for the datasets studied. 

\section{Related Work}

\paragraph{Analysis of NLP Models}  Structural tests for analyzing models' internals include probing \cite{tenney-etal-2019-bert} and attention analysis \cite{jain-wallace-2019-attention, serrano-smith-2019-attention, wiegreffe-pinter-2019-attention, tutek-snajder-2020-staying}. These, along with behavioral tests such as challenge sets \cite{mccoy-etal-2019-right} and checklists \cite{ribeiro-etal-2020-beyond}, are conceptually similar to our experiments, but study different model properties. 

Although gradient-attribution has been extensively studied in NLP, its interplay with free-text rationalization has not.
\citet{wu-mooney-2019-faithful} use feature importance agreement to train the explanation module of a VQA model.
To the best of our knowledge, we are the first to evaluate gradient-attribution reliability for NLP tasks with the ROAR test \cite{hooker2019benchmark}.

\paragraph{Robustness Analysis}

Robustness of post-hoc extractive interpretability methods has been studied \cite{kindermans2019reliability, ghorbani2019interpretation, heo2019fooling, zheng2019analyzing, slack2019can}. 
\citet{zhang2020interpretable} show that saliency maps and model predictions can be independently adversarially attacked in vision and clinical tasks, and conclude this is due to a misalignment between the saliency map generator and model predictor.
Such methods have not been tested for models producing natural language (NL) rationales. Future work could include expanding robustness equivalence (\sect{sec:robustness}) to model discrete edits of input words.

\paragraph{Analyzing Faithfulness}

The aim of our work is to initiate placing models that provide NL rationales on the faithfulness spectrum conceptualized by \citet{jacovi-goldberg-2020-towards}. 
Prior work proposing models \cite{jain-etal-2020-learning, schuff-etal-2020-f1, jacovi2020aligning} and evaluations \cite{deyoung-etal-2020-eraser} of faithful explanation
focus on extractive rationales and generally rely on the sufficiency assumption. \citet{schuff-etal-2020-f1} propose a regularization term to couple answers and extractive explanations on HotPotQA.

Turning to exceptions that focus on natural language rationales, \citet{latcinnik2020explaining} train a differentiable \itor;\irtoo pipeline for \cqa, controlling the complexity of the \irtoo model to increase the likelihood that the model is faithful to the rationale.
\citet{kumar-talukdar-2020-nile} propose an IO$\rightarrow$R;\irtoo pipeline that generates an explanation for every possible NLI label using label-specific explanation generators---an alternative solution to the problem raised in \sect{sec:rationale_quality} for datasets with a small number of shared labels.

\section{Conclusion}

After demonstrating the weaknesses that pipeline models exhibit for free-text rationalization tasks, we propose two measurements of label-rationale association in self-rationalizing models. We find that on three free-text rationalization datasets for CommonsenseQA and SNLI, models based on T5 exhibit high robustness equivalence and feature importance agreement, demonstrating that they pass a necessary sanity check for generating faithful free-text rationales.

Future work can expand analysis to more properties.
We believe this research direction to be important moving forward due to the advantages of large multi-task explanation models, and as a complement to development of interpretable architectures that can be fickle and task-specific. Although our measurements address only necessary and not sufficient properties, by viewing faithful interpretability as a spectrum, we make a step to quantitatively situate common models on it.

\section*{Acknowledgements}
We are grateful to Jonathan Berant, Peter Hase, Alon Jacovi, Yuval Pinter, Mark Riedl, Vered Shwartz, Ian Stewart, Swabha Swayamdipta, and Byron Wallace for feedback on the draft. We thank members of the AllenNLP team at the Allen Institute for Artificial Intelligence (AI2), members of the Entertainment Intelligence lab at Georgia Tech, and reviewers for valuable feedback and discussions. We thank Aaron Chan for pointing out an issue (now corrected) in our use of the term ``simulatability''. This work was done while SW was an intern at AI2.

\bibliography{main,anthology}
\bibliographystyle{acl_natbib}

\appendix

\section{Additional Information}
\label{sec:appendix_info}

\subsection{Overview of Prior Work on Pipelines}
\label{sec:appendixcite}

In \autoref{table:datasets}, we overview the datasets and types of rationales used in prior work on pipelines. The sources of datasets are: CommonsenseQA \cite{talmor-etal-2019-commonsenseqa}, SNLI  \cite{bowman-etal-2015-large}, SST \cite{socher-etal-2013-recursive}, AgNews \cite{del2005ranking}, Evidence Inference \cite{lehman-etal-2019-inferring}, Movie Reviews \cite{zaidan-eisner-2008-modeling}, MultiRC \cite{khashabi-etal-2018-looking}, LGD \cite{linzen2016assessing}, 20 News \cite{lang1995newsweeder}, Amazon Reviews \cite{mcauley2013hidden}, Beer Reviews \cite{mcauley2012learning}, BoolQ \cite{clark-etal-2019-boolq}, FEVER \cite{thorne-etal-2018-fever}.

\subsection{Details of Datasets}
\label{sec:appendix_dataset}

We summarize dataset statistics in \autoref{table:stats}. The two versions (v1.0, v1.11) of CoS-E correspond to the first and second versions of the CommonsenseQA dataset. CoS-E v1.11 has some noise in its annotations \cite{narang2020wt5}.\footnote{\url{https://github.com/salesforce/cos-e/issues/2}} This is our primary motivation for reporting on v1.0 as well, which we observe does not have these issues.

\subsection{Details of T5}
\label{sec:appendix_t5_details}
The T5 model \cite{raffel2019exploring} is pretrained on a multi-task
mixture of unsupervised and supervised tasks, including machine translation, question answering, abstractive summarization, and text classification. Its inputs and outputs to every task are text sequences; we provide the input-output formatting for training and decoding of our T5 models in~\autoref{tab:t5_data_format}. T5 can provide any word in the vocabulary as an answer.

\subsection{Implementation Details}
\label{sec:hparams}

We use Huggingface Datasets\footnote{\url{https://huggingface.co/docs/datasets/master/\#}} to access all datasets, and Huggingface Transformers \cite{wolf-etal-2020-transformers} to access pretrained T5 weights and tokenizer.
To optimize, we use Adam with $\epsilon = 1$e-8, $\beta_1 = 0.9$, and $\beta_2 = 0.99$. We use gradient clipping to a maximum norm of 1.0 and a dropout rate of 0.1. We train each model on a NVIDIA RTX 8000 GPU (48GB memory) for maximum 200 epochs with a batch size of 64 and a learning rate linearly decaying from 5e-5. Training ends if the validation set loss has not decreased for 10 epochs. Early stopping occurs within 15 epochs for most models. Most CoS-E models train in less than 1 hour and most E-SNLI models in around 30. At inference time, we greedy-decode until an EOS token is generated (or for 200 tokens). Approximating the 64-batch model with a batch-size of 16 and 4 gradient accumulation steps on 8GB memory cloud GPUs, we sweep starting learning rates of 1e-2, 1e-3, 1e-4, 5e-5, and 1e-5. The two largest learning rates never result in good performance. Among the smallest three rates, performance across all model variants (\itor, \itoor, \rtoo, I$\rightarrow$O, and \irtoo) on E-SNLI and CoS-E v1.0 never varies by more than 1.58\% accuracy or 0.34 BLEU. 

\subsection{Note on Robustness Equivalence Convergence}
\label{ssec:convergence}

Worst-case model performance under large noise values in the Robustness Equivalence experiments (Figures \ref{fig:robustness} and \ref{fig:cose-v2-robustness}) reaches 0 rather than random accuracy due to structure of the models' output. The \itoor model is trained to produce a delimiter to distinguish the label from the rationale in a long string of output tokens. When it fails to produce the delimiter under high noise, we cannot delineate the label from the rationale in the output where multiple answer choices are often mentioned, so we mark the label as incorrect.

\subsection{Further Discussion of Rationale Quality Metric}
\label{ssec:caveats}

Traditional simulatability is often considered to be lower-bounded at 0, assuming model-predicted explanations are consistent with model-predicted labels, because a model-predicted explanation should not provide negative utility when given as input to a simulator predicting that label, unless the explanation is inconsistent (i.e., it explains a different label). 
Our rationale quality metric does not have this property, since model-predicted explanations that explain an incorrect label may provide negative utility toward predicting the gold label. As reported in the paper, it is commonly negative in our experiments.

The limitation of our rationale quality metric is that low scores may be due to poor-quality rationales (what we aim to measure) or poor label prediction performance of the model generating the rationales, assuming consistent predicted labels and rationales. Future work may focus on a robust version of simulatability that can both separate these confounders \emph{and} be computed when model predictions are noisy or ill-defined.

\section{Additional Results}
\label{sec:appendix_results}

\begin{figure*}[b]
\centering
\includegraphics[width=\columnwidth]{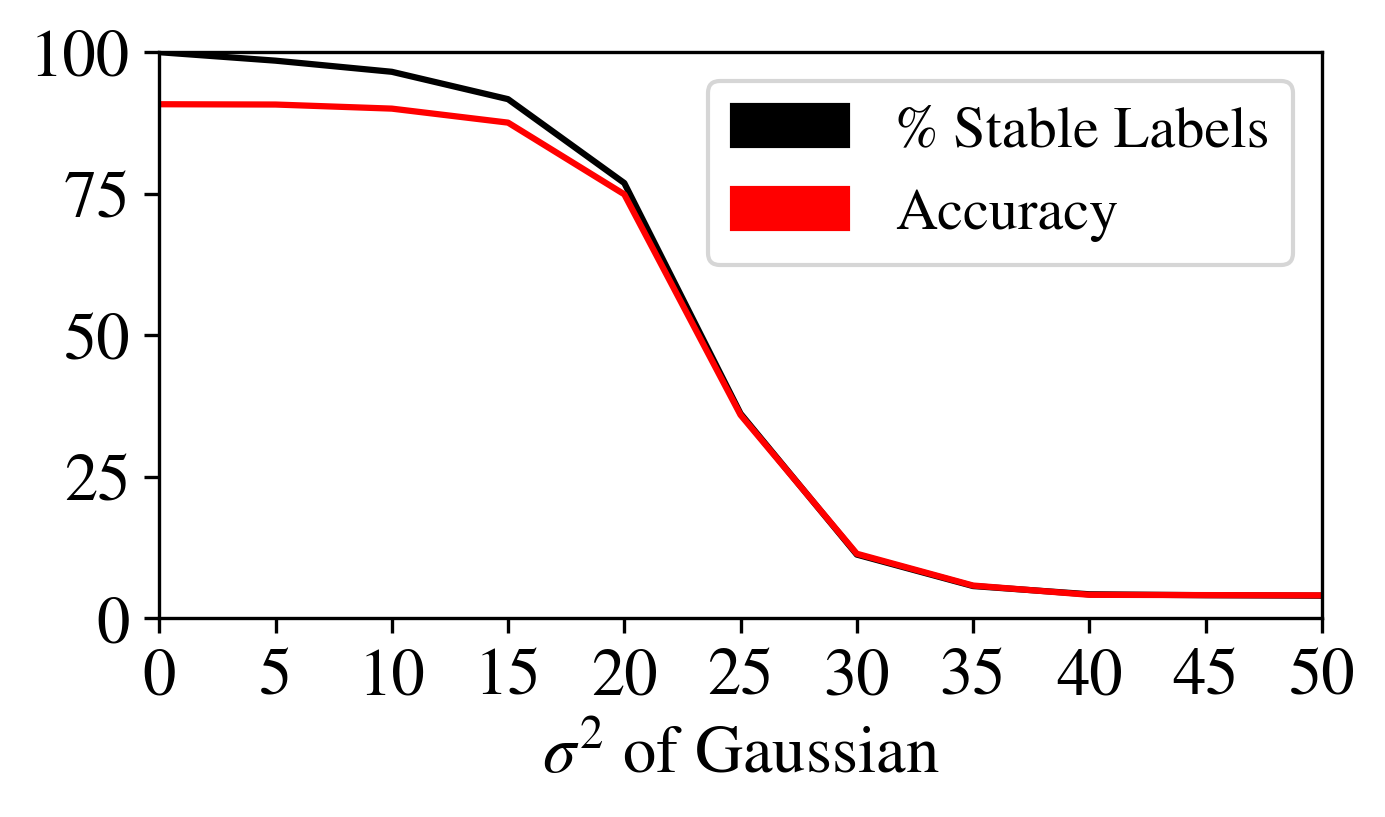}
 \includegraphics[width=\columnwidth]{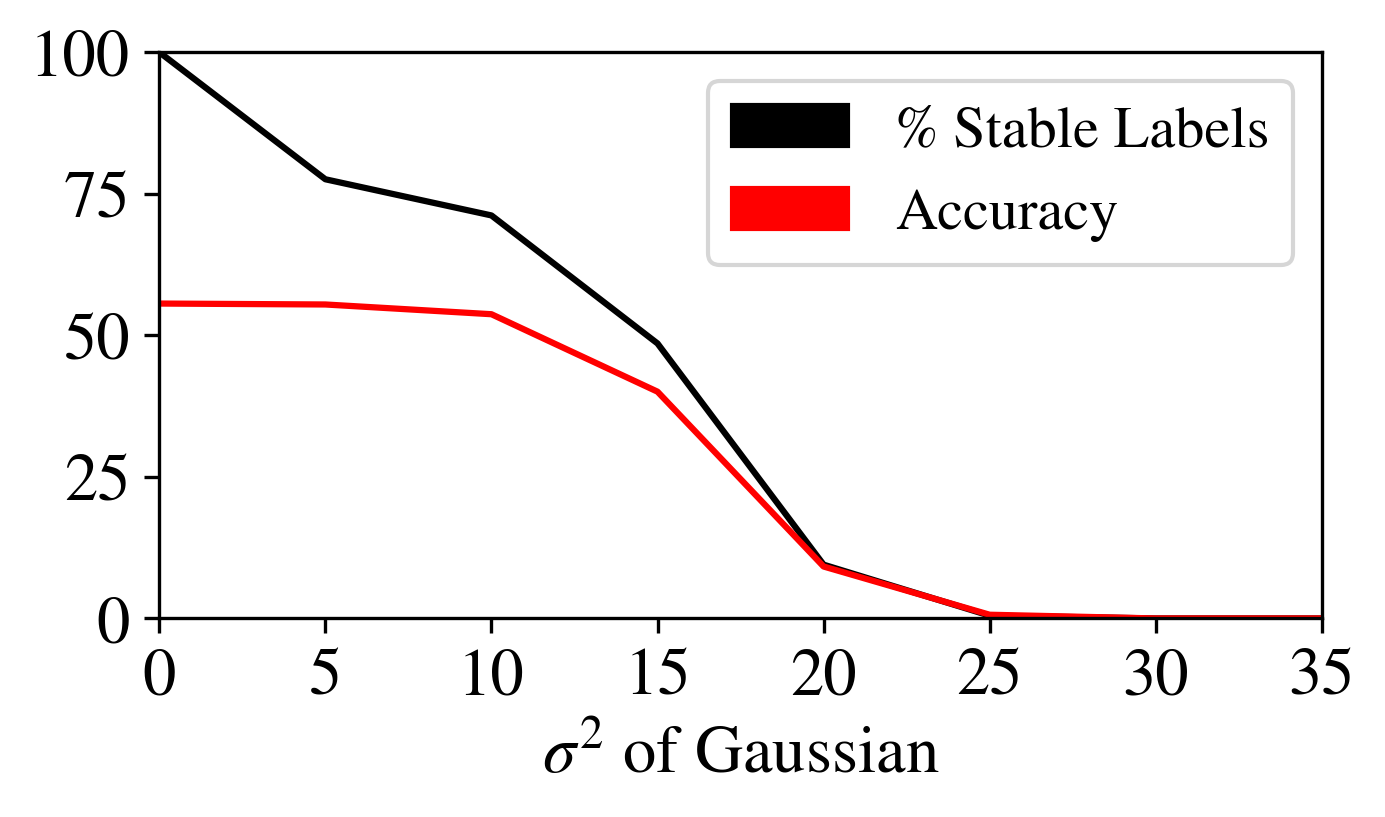}
\caption{
Results of the label portion of the robustness equivalence test for E-SNLI (left) and CoS-E v1.11 (right). Accuracy of the \itoor model (red) and \% stable labels in the \itoor model (black) show that most changes take place in the 10-20 $\sigma^2$ range for CoS-E and 15-30 $\sigma^2$ for E-SNLI. See \sect{sec:robustness}.}
\label{fig:cose-v2-robustness}
\end{figure*}

\begin{figure*}[t]
\centering
\includegraphics[width=0.5\linewidth]{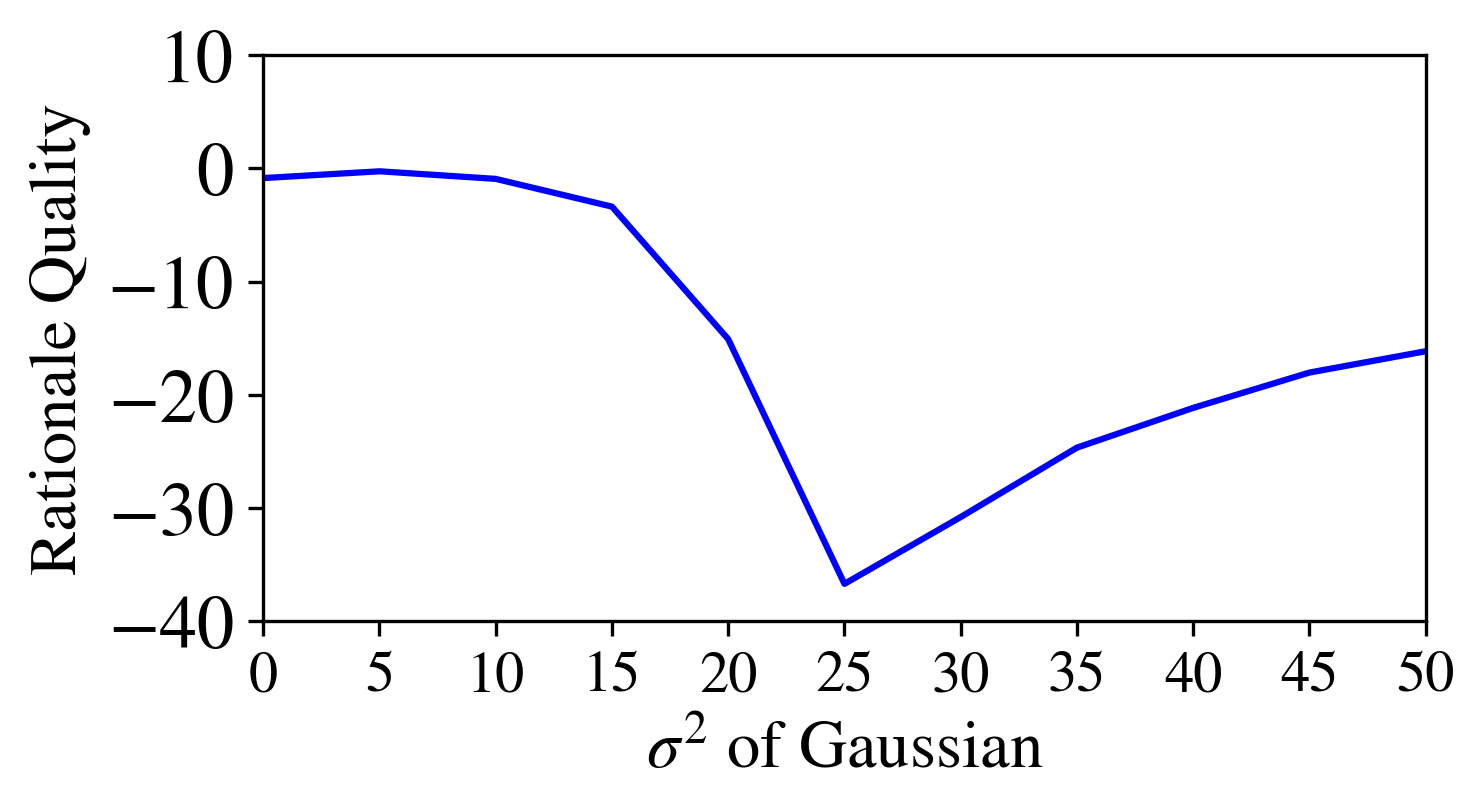}
\caption{Results of the rationale portion of the robustness equivalence test for E-SNLI.
}
\label{fig:robustness_label_flip_appendix}
\end{figure*}

\begin{figure*}[t]
  \centering
  \includegraphics[width=1.2\columnwidth]{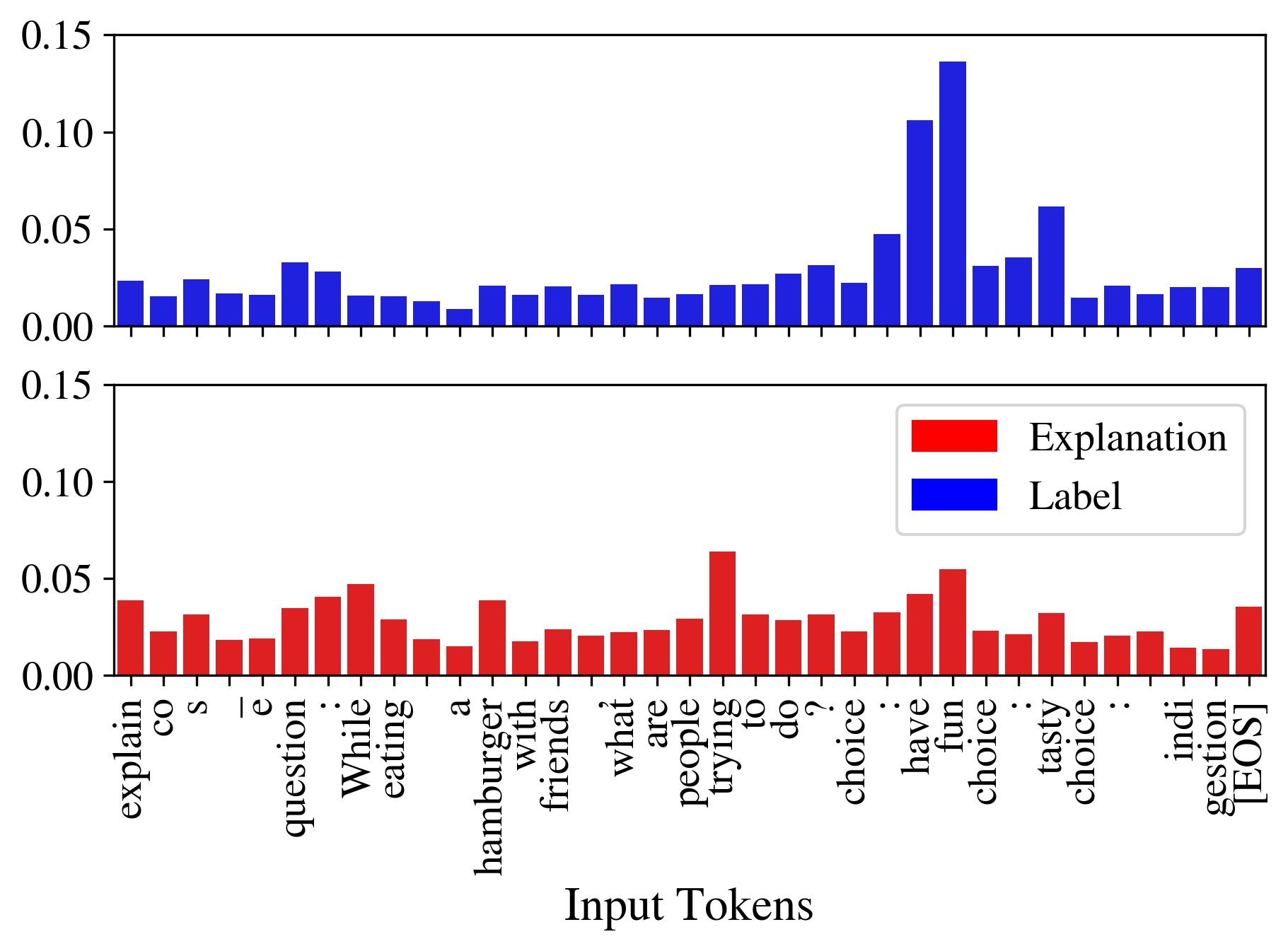}
  \caption{Normalized attributions for the running CoS-E v1.0 example in Figures \ref{fig:pipeline}--\ref{fig:joint}. The decoded label is ``have fun'' and generated rationale ``having fun is the only thing that people are trying to do''. Important input terms vary across the two loss terms. For example, the predicted label term assigns high importance to the predicted answer choice, ``have fun'', while the explanation attends more uniformly over the input with peaks on relevant entities and verbs such as ``trying''. In this example, the explanation- and label-attribution vectors are each $L_1$-normalized in order to compare the relative importance of tokens (irrespective of gradient magnitudes).
  } 
  \label{fig:gradients}
\end{figure*}

\begin{figure*}[b]
\centering
   \includegraphics[width=0.6\linewidth]{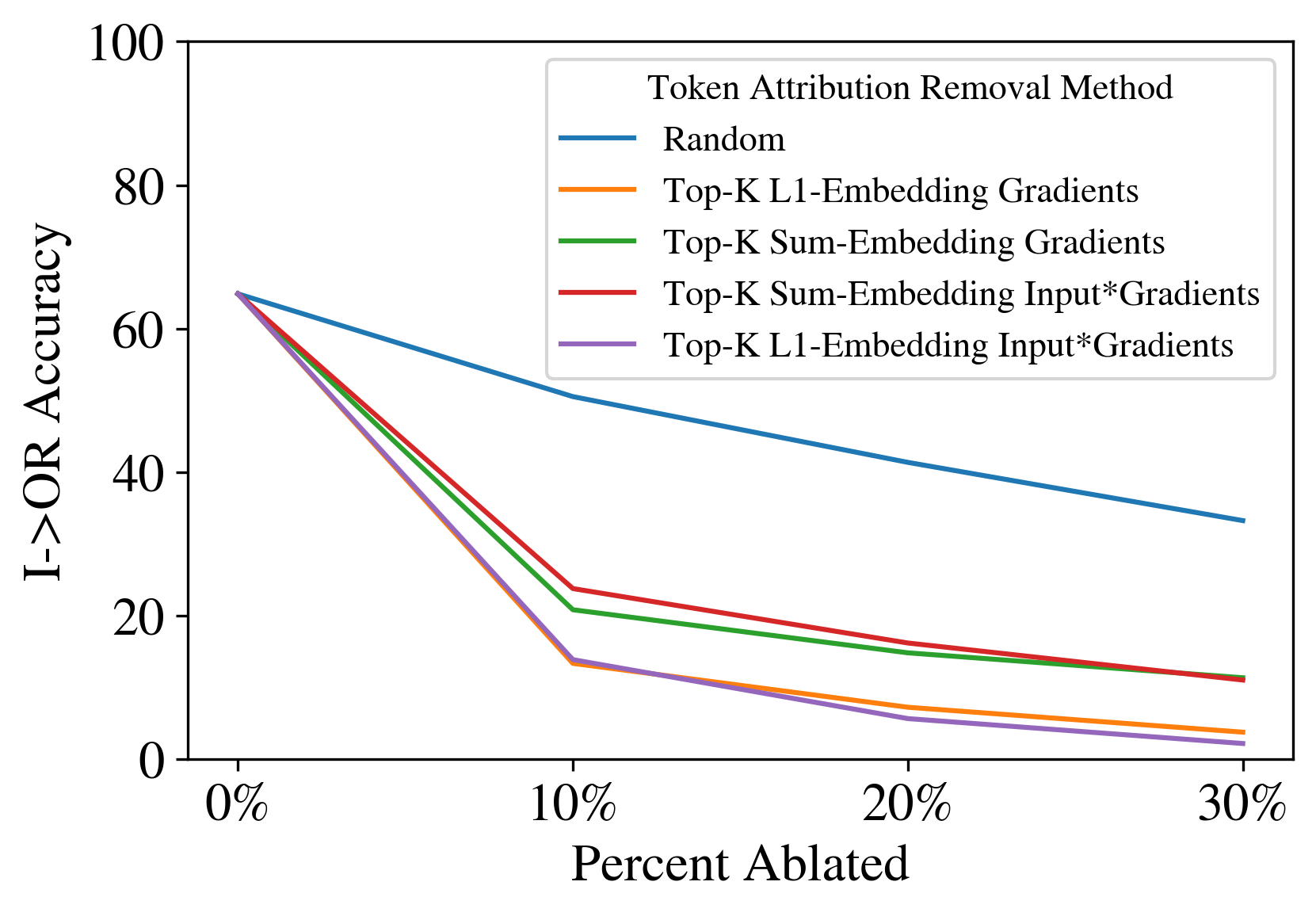}
\caption{Effect of various gradient attribution methods on the ROAR test at $k=10-30\%$ occlusion for the CoS-E v1.0 validation set. We compute attributions with respect to the label logit and measure label accuracy of the resulting model after masking and re-training (see \sect{sec:gradients} for details). The largest drop in performance comes from the $L_1$ norm embedding-combination method, and raw gradients are not significantly different from input*gradient. On average, input*gradient and raw gradients share 17\% of tokens in the top 10\%, 24\% of tokens in the top 20\%, and 31\% of tokens in the top 30\%.}
\label{fig:roar_testing_methods}
\end{figure*}

\begin{figure*}[t]
\centering
\begin{subfigure}[t]{\textwidth}
\centering
   \includegraphics[width=0.48\columnwidth]{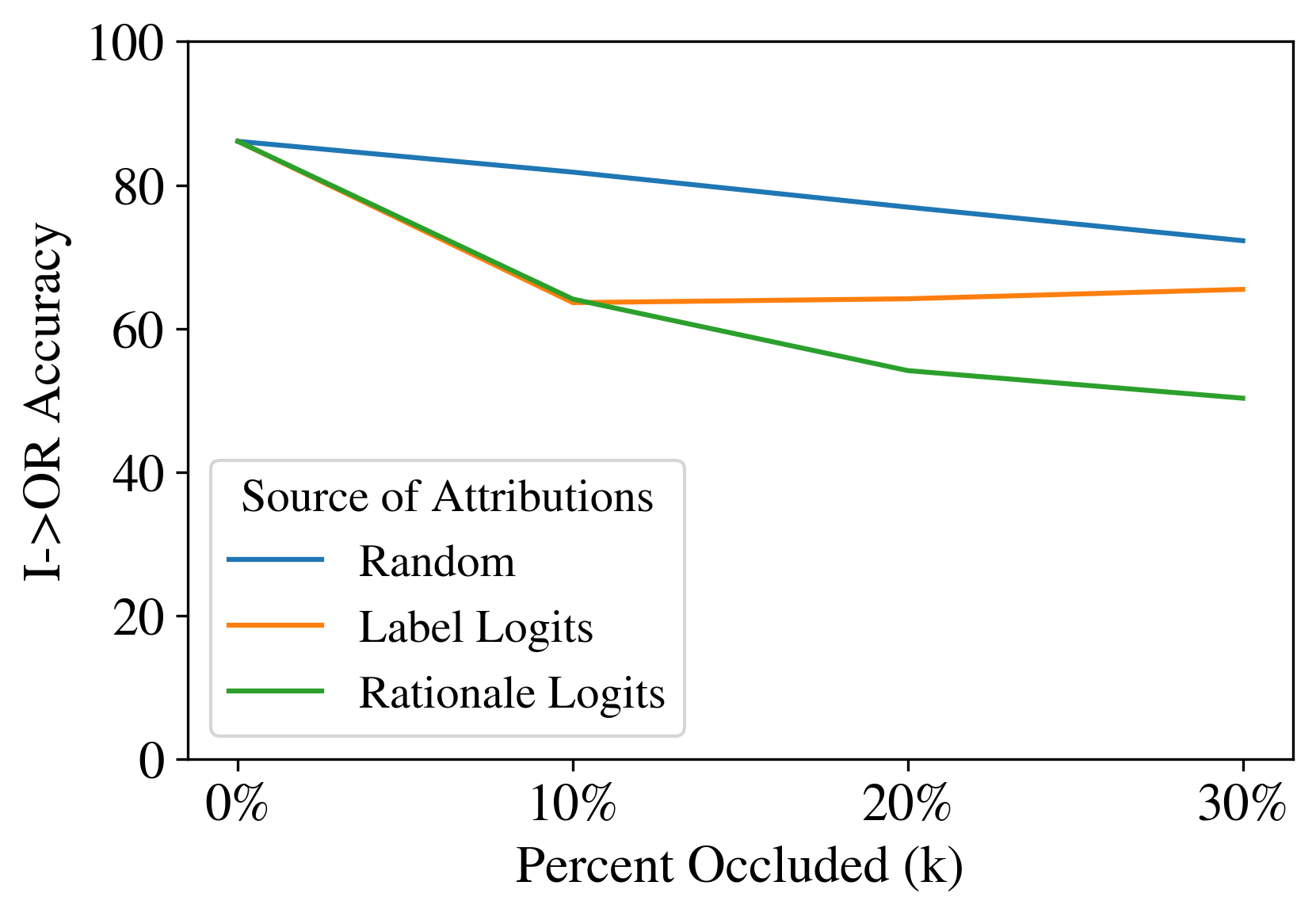}
   \includegraphics[width=0.48\columnwidth]{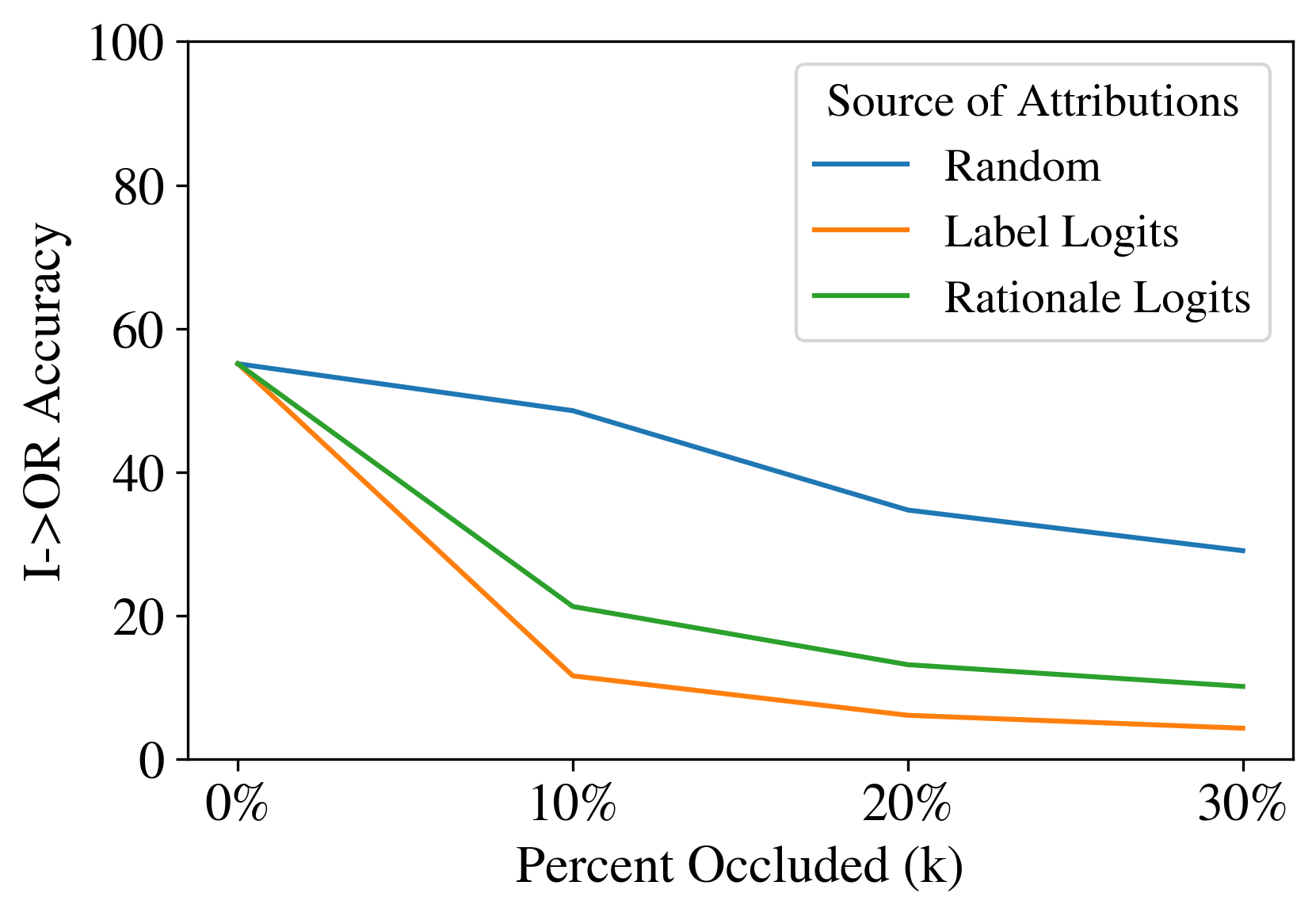}
\caption{Impact of occlusion by source of attribution on label accuracy.}
\label{part-a}
\end{subfigure}\text{ }
\begin{subfigure}[t]{\textwidth}
\centering
   \includegraphics[width=0.48\columnwidth]{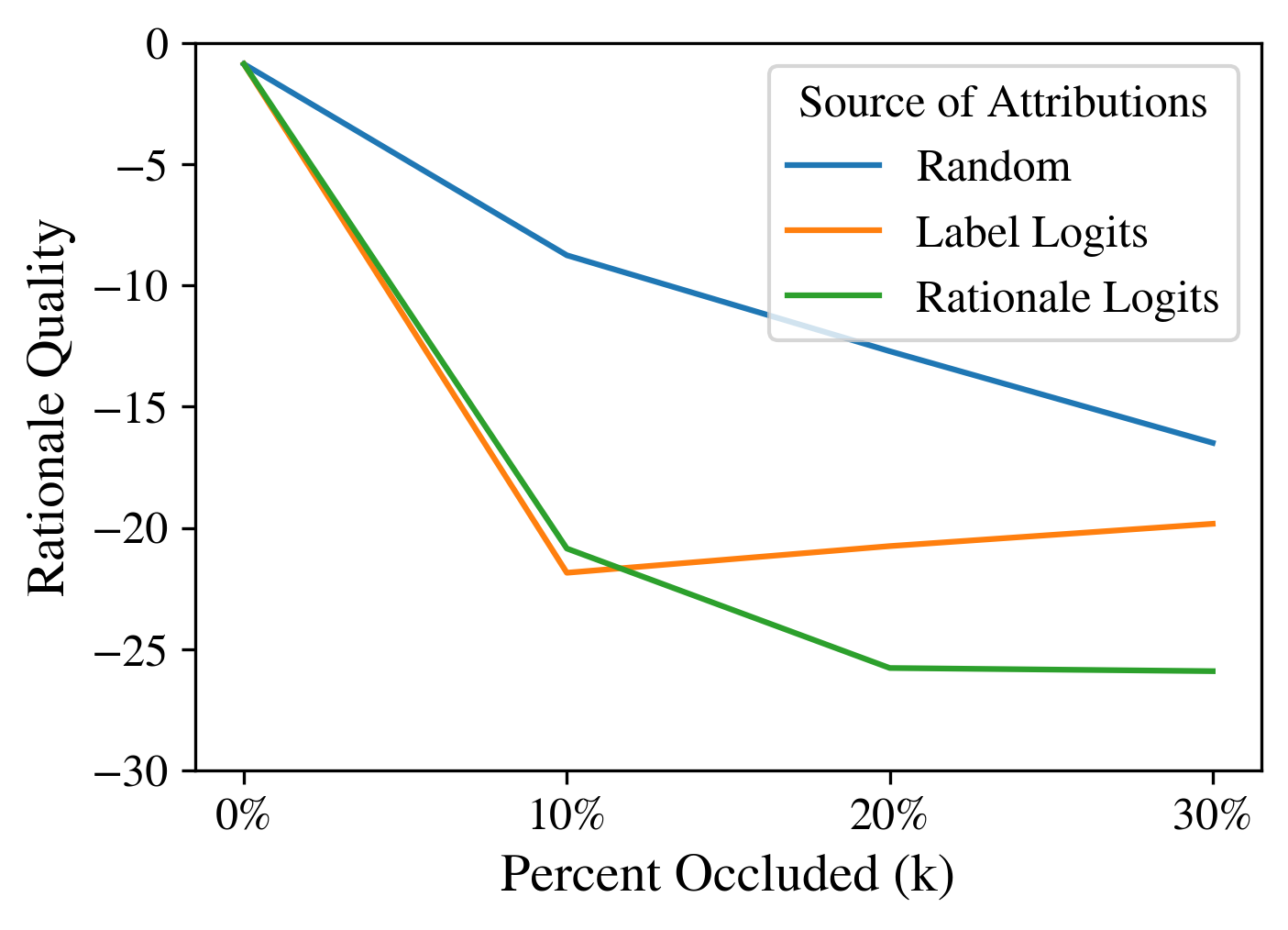}
   \includegraphics[width=0.48\columnwidth]{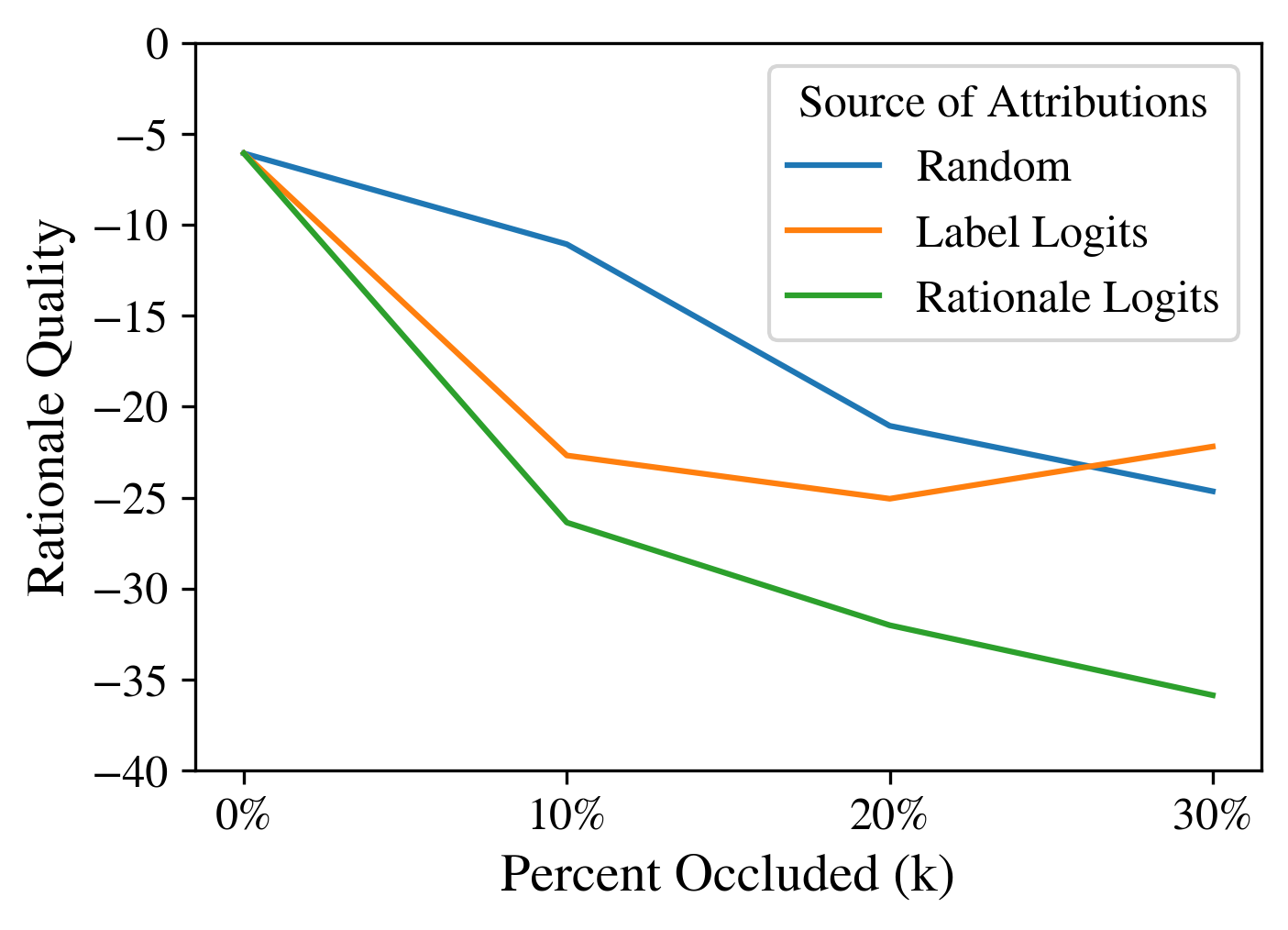}
\caption{Impact of occlusion by source of attribution on rationale quality.}
\label{part-b}
\end{subfigure}
\caption{ROAR Feature Importance Agreement results on E-SNLI (left) 
and CoS-E v1.11 (right). \autoref{part-a} shows label accuracy of the \itoor model. \autoref{part-b} shows quality of generated rationales from the \itoor model. 
}
\label{fig:supp_roar_results}
\end{figure*}

We provide additional results that supplement the main body of the paper:
\begin{compactitem}
\item \autoref{table:baselines} presents results comparing the self-rationalizing T5 model to baselines.
\item \autoref{table:pipelines} presents results comparing the self-rationalizing T5 model to its pipeline variant (from  \sect{sec:self_rationalizing}).
\item \autoref{fig:cose-v2-robustness} presents robustness equivalence label results for E-SNLI and CoS-E v1.11 (\sect{sec:robustness}). 
\item \autoref{fig:robustness_label_flip_appendix} presents robustness equivalence rationale results for E-SNLI (\sect{sec:robustness}). CoS-E v1.0 and v1.11 rationale results are included in \autoref{fig:robustness_label_flip} in main paper.
\item \autoref{fig:gradients} presents an example of $L_1$-normalized gradient attributions for a single instance (\sect{sec:gradients}).
\item \autoref{fig:roar_testing_methods} presents a comparison of attribution methods on the ROAR reliability check (\sect{sec:gradients}) for CoS-E v1.0.
\item \autoref{fig:supp_roar_results} presents results of the ROAR feature importance agreement measure (\sect{sec:gradients}) for CoS-E v1.11 and E-SNLI.
\item Tables \ref{table:cose_noise_1}-\ref{table:cose_noise_3} contain additional validation set examples (non-cherry-picked) of noised inputs for CoS-E v1.0 (\sect{sec:robustness}).
\end{compactitem}

\begin{table*}
    \centering
    \resizebox{\textwidth}{!}{
    \begin{tabular}{lcccccc}
    \toprule
        \textbf{Dataset} & \textbf{Num. Instances} & \textbf{Input Length} & \multicolumn{2}{c}{\textbf{Extractive Rationale}} & \multicolumn{2}{c}{\textbf{Natural Language Rationale}}\\ 
        \cmidrule(lr){2-2}\cmidrule(lr){3-3}\cmidrule(lr){4-5}\cmidrule(lr){6-7}
        & Train-Val-Test & \# Tokens & \# Tokens & \% of doc. & \# Tokens & \% of doc. \\
        \midrule
        E-SNLI & 549,367-9,842-9,824 & 20.27 +/-- 6.95 & 4.01 +/-- 3.01 & 21.30 +/-- 15.82 &  12.39 +/-- 6.43 & 65.67 +/-- 35.46\\ 
        CoS-E v1.0 & 7,610-950-none & 13.69 +/-- 5.97 & 4.57 +/-- 4.16 & 35.36 +/-- 27.22 & 12.74 +/-- 6.99 & 108.18 +/-- 77.26\\
        CoS-E v1.11 & 9,741-1,221-none & 13.40 +/-- 5.77 & 6.80 +/-- 5.79 & 53.24 +/-- 36.05 & 6.97 +/-- 4.14 & 58.01 +/-- 39.60\\
        \bottomrule
    \end{tabular}
    }
    \caption{Statistics on datasets with ground-truth rationales. Results are presented as mean (one standard deviation) on the training set. CoS-E does not contain test set rationale annotations.}  
    \label{table:stats}
\end{table*}
\begin{table*}
\begin{tabular}{p{\textwidth}}
\toprule
\textbf{The \itoor and \itor;\rtoo rationale generator's inputs}:\\
explain cos\textunderscore e question: [question] choice: [choice\textunderscore0] choice: [choice\textunderscore1] choice: [choice\textunderscore2] \\ 
explain nli hypothesis: [hypothesis] premise: [premise].\\
\midrule
\textbf{The \itor;\rtoo pipeline label predictor's input:}\\
cos\textunderscore e choice: [choice\textunderscore0] choice: [choice\textunderscore 1] choice: [choice\textunderscore 2] explanation: [free-text rationale] \\
nli explanation: [free-text rationale]\\
\midrule
\textbf{The \itoor models' outputs are trained and decoded as:}\\
\text{}[label] explanation: [free-text rationale]\\
\bottomrule
\end{tabular}
\caption{T5 input-output formatting.}
\label{tab:t5_data_format}
\end{table*}
\begin{table*}
    \centering    
    \resizebox{\textwidth}{!}{
    \begin{tabular}{|l|r|r|r|r|r|}
    \toprule
        & \textbf{Random} &  \textbf{Majority-Vote} &  & \\
        \textbf{Dataset} & \textbf{Accuracy} &  \textbf{Accuracy} & \textbf{I$\rightarrow$O (T5)} & \textbf{\itoor} & $\Delta$ \\
        \midrule
        E-SNLI & 33.33 & 33.39 & 90.95 (84.01\textsuperscript{$\ddagger$}) & 90.81 (83.96\textsuperscript{$\ddagger$},  90.9\textsuperscript{$\dagger$}) & --0.14 (--0.05\textsuperscript{$\ddagger$})\\
        CoS-E v1.0 & 33.33 & 33.75 & 69.16 (63.8\textsuperscript{$\mathsection$}) & 64.84  & --4.32\\
        CoS-E v1.11 & 20.0 & 20.31 & 61.75 & 55.61 (59.4\textsuperscript{$\dagger$}) & --6.14\\
        \bottomrule
    \end{tabular}}
    \caption{Label accuracy on baseline I$\rightarrow$O T5 models versus their rationalizing \itoor variants fine-tuned for each dataset. We observe that adding rationalization results in some loss in accuracy. We also validate that T5-Base models outperform other architectures.
    Source of prior results in parentheses: $\dagger$  \citet{narang2020wt5} using T5, $\ddagger$ \citet{camburu2018snli} using a bi-directional LSTM, and $\mathsection$ \citet{rajani-etal-2019-explain} using BERT.}
    \label{table:baselines}
\end{table*}
\begin{table*}
    \centering
    \begin{tabular}{|l|r|r|r|}
    \toprule
        \textbf{Dataset} &  \textbf{\itoor} & \textbf{\itor;\rtoo} & $\Delta$\\ 
        \midrule
        E-SNLI & 90.81 (83.96\textsuperscript{$\ddagger$}) &  89.11 (81.71\textsuperscript{$\ddagger$}) &  --1.70 (--2.25\textsuperscript{$\ddagger$})\\
        CoS-E v1.0 & 64.84 & 53.47 & --11.37 \\
        CoS-E v1.11 & 55.61 & 45.45 & --10.16 \\
        \bottomrule
    \end{tabular}
    \caption{Label accuracy on the joint self-rationalizing model \itoor compared to a pipeline using natural language rationales. We observe that \itoor models have stronger task performance. Source of prior results in parentheses: $\ddagger$ \citet{camburu2018snli} using bi-directional LSTMs.}
    \label{table:pipelines}
\end{table*}


\begin{table*}
    \centering
    \resizebox{\textwidth}{!}{
    \begin{tabular}{l|l}
    \toprule
        \textbf{$\sigma^2$} & \textbf{Predicted Output}\\ 
        \midrule
        0 & house \textbf{explanation:} a house is the only place that would have air conditioning.\\
        5 & house \textbf{explanation:} a house is the only place that would have air conditioning.\\
        10 & house \textbf{explanation:} a house is the only place that would have air conditioning.\\
        15 & <extra\_id\_0> house \textbf{explanation:} a house is the only place that will have air conditioning.\\ 
        20 & <extra\_id\_0> movie theatre \textbf{explanation:} movie theatre is the only option that is not a movie. 911 911 911\dots\\
        25 & <extra\_id\_0> explain<extra\_id\_1> explain<extra\_id\_2> explain<extra\_id\_3> explain<extra\_id\_4> explain<extra\_id\_5> movie theatre<extra\_id\_6>\dots\\
        30 & house of house of house of house of house of house of house of house of house of house of house of house of house \dots\\ 
        35 & house of house of office office office office office office office office office office office office office office office office office office\dots\\
        \bottomrule
    \end{tabular}
    }
    \caption{Noised output of the \itoor model for the CoS-E v1.0 example ``A man wants air conditioning while we watches the game on Saturday, where will it likely be installed?'' The correct answer is ``house''.}  
    \label{table:cose_noise_1}
\end{table*}

\begin{table*}
    \centering
    \resizebox{\textwidth}{!}{
    \begin{tabular}{l|l}
    \toprule
        \textbf{$\sigma^2$} & \textbf{Predicted Output}\\ 
        \midrule
        0 & stress \textbf{explanation:} a computer is used to communicate with a granddaughter.\\
        5 & stress \textbf{explanation:} a computer is used to communicate with a granddaughter.\\
        10 & stress \textbf{explanation:} a computer is used to talk to people.\\
        15 & stress \textbf{explanation:} a computer is used to talk to people.\\
        20 & <extra\_id\_0> is using a computer to<extra\_id\_1> to talk to<extra\_id\_2> is using a computer to talk to a person is using a computer to talk to a person\dots\\
        25 & <extra\_id\_0> answer: answer: answer: answer: answer: answer: answer: answer: answer: answer: answer: answer: answer:\dots\\
        30 &  <extra\_id\_0> answer: answer: answer: answer: answer: answer: answer: answer: answer: answer: answer: answer: answer:\dots\\ 
        35 & office of the office of the office of the office of the office of the office of the office of the office of the office of the office of the office \dots \\
        \bottomrule
    \end{tabular}
    }
    \caption{Noised output of the \itoor model for the CoS-E v1.0 example ``If a person is using a computer to talk to their granddaughter, what might the computer cause for them?'' The correct answer is ``happiness''.}  
    \label{table:cose_noise_2}
\end{table*}

\begin{table*}
    \centering
    \resizebox{\textwidth}{!}{
    \begin{tabular}{l|l}
    \toprule
        \textbf{$\sigma^2$} & \textbf{Predicted Output}\\ 
        \midrule
        0 & transfer of information \textbf{explanation:} transfer of information is the only option that would be appropriate when communicating with a boss. \\
        5 & transfer of information \textbf{explanation:} transfer of information is the only option that would be appropriate when communicating with a boss. \\
        10 & transfer of information \textbf{explanation:} transfer of information is the only option that would be appropriate when communicating with a boss. \\
        15 & transfer of information \textbf{explanation:} transfer of information is the only option that would be appropriate when communicating with my boss. \\
        20 & transfer of information: transfer of information is the only thing that is transfer of information. transfer of information is the only thing that is\dots\\
        25 & transfer of information: transfer of information is information. transfer of information is information. transfer of information is information.\dots\\
        30 & i believe that is the answer of the question. argument is the answer of the question. argument is the answer of the question. argument is the answer of the question.\dots\\ 
        35 & i can't handle the argument argument argument is the answer of the argument argument is the answer of the argument\dots\\
        \bottomrule
    \end{tabular}
    }
    \caption{Noised output of the \itoor model for the CoS-E v1.0 example ``When communicating with my boss, what should I do?''. The correct answer is ``transfer of information''.}  
    \label{table:cose_noise_3}
\end{table*}

\end{document}